%% file: dynavol.tex
\documentclass[10pt,journal,compsoc]{IEEEtran}
\usepackage{ifpdf}
\ifpdf
\else
\fi
\ifCLASSINFOpdf
   \usepackage{graphicx}
   \graphicspath{{../pdf/}{../jpeg/}}
   \DeclareGraphicsExtensions{.pdf,.jpeg,.png}
\else
   \usepackage[dvips]{graphicx}
   \graphicspath{{../eps/}}
   \DeclareGraphicsExtensions{.eps}
\fi
\usepackage{amsmath}
\usepackage{array}

\ifCLASSOPTIONcompsoc
 \usepackage[caption=false,font=footnotesize,labelfont=sf,textfont=sf]{subfig}
\else
 \usepackage[caption=false,font=footnotesize]{subfig}
\fi

\usepackage{stfloats}
\ifCLASSOPTIONcaptionsoff
 \usepackage[nomarkers]{endfloat}
\let\MYoriglatexcaption\caption
\renewcommand{\caption}[2][\relax]{\MYoriglatexcaption[#2]{#2}}
\fi

\usepackage{algorithm}
\usepackage{algorithmic}

\let\MYorigsubfloat\subfloat
\renewcommand{\subfloat}[2][\relax]{\MYorigsubfloat[]{#2}}
\usepackage{xspace}
\makeatletter
\DeclareRobustCommand\onedot{\futurelet\@let@token\@onedot}
\def\@onedot{\ifx\@let@token.\else.\null\fi\xspace}

\makeatother

\usepackage{enumitem}
\usepackage{caption}
\usepackage{times}
\usepackage{epsfig}
\usepackage{graphicx}
\usepackage{amsmath}
\usepackage{amssymb}
\usepackage{eso-pic}
\usepackage{helvet}
\usepackage{courier}
\usepackage{url}
\usepackage{mathrsfs}
\usepackage{multirow}
\usepackage{comment}
\usepackage{marvosym}
\usepackage{bbm}
\usepackage{soul}
\usepackage[utf8]{inputenc}
\usepackage{amsfonts}
\usepackage{nicefrac}
\usepackage{microtype}
\usepackage{booktabs} 
\usepackage{algorithm}
\usepackage{makecell}
\usepackage{multirow}

\usepackage{bbding}
\usepackage[numbers]{natbib}
\usepackage[breaklinks=true,bookmarks=false,colorlinks]{hyperref}
\usepackage{cleveref}
\usepackage{setspace}

\usepackage[normalem]{ulem}

\newcommand{\myparagraph}[1]{\vspace{5pt} \noindent \textbf{#1}}

\newcommand{\revise}[1]{{#1}}

\definecolor{MyDarkBlue}{rgb}{0,0.5,1}
\definecolor{MyDarkGreen}{rgb}{0.02,0.6,0.02}
\definecolor{MyDarkRed}{rgb}{0.8,0.02,0.02}
\definecolor{MyDarkOrange}{rgb}{0.40,0.2,0.02}
\definecolor{MyPurple}{rgb}{111,0,255}
\definecolor{MyRed}{rgb}{1.0,0.0,0.0}
\definecolor{MyGold}{rgb}{0.75,0.6,0.12}
\definecolor{MyDarkgray}{rgb}{0.66, 0.66, 0.66}
\newcommand{\model}{DynaVol}
\newcommand{\newModel}{DynaVol-S}

\newcommand{\major}[1]{{#1}}

\begin{document}



\title{Dynamic Scene Understanding through Object-Centric Voxelization and Neural Rendering}

%
%
%
%

\author{Yanpeng~Zhao,
        Yiwei~Hao,
        Siyu~Gao,
        Yunbo~Wang,
        Xiaokang~Yang,~\IEEEmembership{Fellow,~IEEE}
\IEEEcompsocitemizethanks{
\IEEEcompsocthanksitem The authors are with MoE Key Lab of Artificial Intelligence, AI Institute, Shanghai Jiao Tong University, China.
\IEEEcompsocthanksitem Corresponding author: Y. Wang, yunbow@sjtu.edu.cn.
\IEEEcompsocthanksitem Code: \url{https://github.com/zyp123494/DynaVol}.
}
}

\markboth{IEEE Transactions on Pattern Analysis and Machine Intelligence,~EARLY ACCESS, February~2025}%
{Zhao \MakeLowercase{\textit{et al.}}: DynaVol-S: Unsupervised Learning for Dynamic
Scenes through Object-Centric Voxelization}

\IEEEtitleabstractindextext{%

\input{text/abstract}

}

\maketitle

\IEEEdisplaynontitleabstractindextext

%
\IEEEpeerreviewmaketitle

\input{text/intro}
\input{text/problem_setup}
\input{text/method.tex}

\input{text/expri.tex}

\input{text/related.tex}
\input{text/concl.tex}


%



\ifCLASSOPTIONcompsoc
  \section*{Acknowledgments}
\else
  \section*{Acknowledgment}
\fi

This work was supported by the National Natural Science Foundation of China (Grant No. 62250062, 62106144), the Shanghai Municipal Science and Technology Major Project (Grant No. 2021SHZDZX0102), and the Fundamental Research Funds for the Central Universities.


\ifCLASSOPTIONcaptionsoff
  \newpage
\fi



%

\bibliographystyle{IEEEtran}
\bibliography{IEEEabrv,dynavol.bib}

\input{dynavol_suppl}

\end{document}

%% file: text/abstract.tex
\begin{abstract}

Learning object-centric representations from unsupervised videos is challenging. Unlike most previous approaches that focus on decomposing 2D images, we present a 3D generative model named DynaVol-S for dynamic scenes that enables object-centric learning within a differentiable volume rendering framework. The key idea is to perform object-centric voxelization to capture the 3D nature of the scene, which infers per-object occupancy probabilities at individual spatial locations. These voxel features evolve through a canonical-space deformation function and are optimized in an inverse rendering pipeline with a compositional NeRF. Additionally, our approach integrates 2D semantic features to create 3D semantic grids, representing the scene through multiple disentangled voxel grids. DynaVol-S significantly outperforms existing models in both novel view synthesis and unsupervised decomposition tasks for dynamic scenes. By jointly considering geometric structures and semantic features, it effectively addresses challenging real-world scenarios involving complex object interactions. Furthermore, once trained, the explicitly meaningful voxel features enable additional capabilities that 2D scene decomposition methods cannot achieve, such as novel scene generation through editing geometric shapes or manipulating the motion trajectories of objects.

\end{abstract}

%% file: text/intro.tex
\section{Introduction}

Unsupervised learning of the physical world is of great importance but challenging due to the intricate entanglement between the spatial and temporal information~\citep{wu2015galileo,santoro2017simple,greff2020binding}.
%
%
Existing approaches primarily leverage the consistency of the dynamic information across consecutive video frames but tend to ignore the 3D nature, resulting in a multi-view mismatch of 2D object segmentation~\citep{kabra2021simone,Elsayed2022SAViTE,singh2022simple}.
This paper explores a novel research problem: unsupervised 3D dynamic scene decomposition. Unlike previous 2D approaches, our method naturally ensures 3D-consistent scene understanding and achieves an explicit knowledge of the object geometries and physical interactions in the dynamic scenes.

We propose \newModel{}, an inverse graphics model based on neural radiance field (NeRF), \revise{to learn 3D scene layouts from sequential video frames captured by a moving monocular camera.}
However, the classic canonical-space rendering pipeline does not support object-centric learning. We identify two key challenges: First, how to decouple the underlying dynamics of each object from its visual appearance and capture the \textit{local geometric structures} of each object. Second, how to integrate the \textit{global semantic priors} of the scene, typically extracted from 2D foundation models, into 3D volume rendering, which is crucial for inferring 3D geometries given partial observations, complex real-world structures, and severe occlusions.

\begin{figure*}[t]
\centering
\centerline{\includegraphics[width=0.95\linewidth]{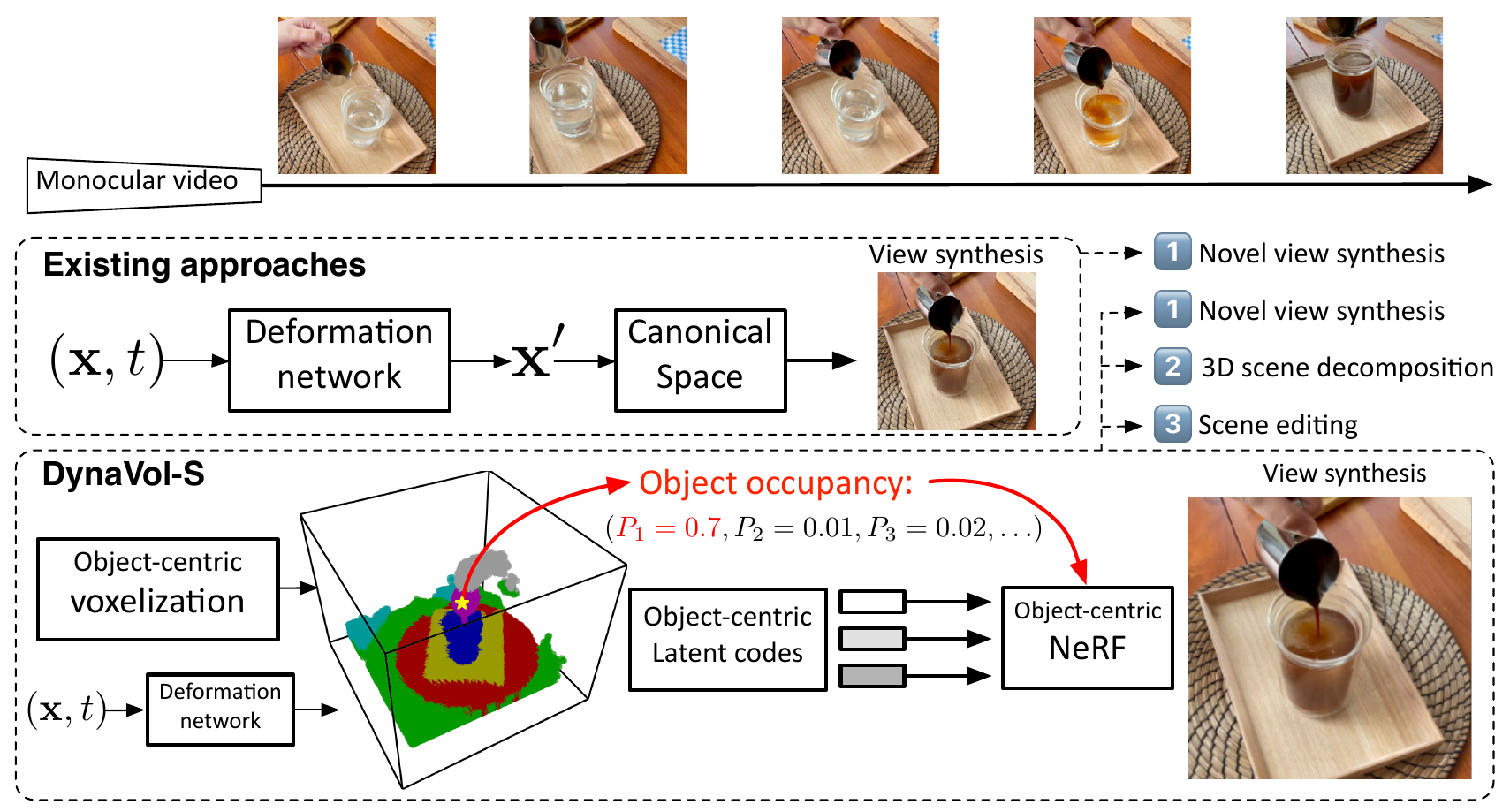}}
\vspace{-5pt}
\caption{
\newModel{} explores unsupervised object-centric decomposition in 3D dynamic scenes within an inverse rendering framework. Unlike previous canonical-space neural rendering techniques, such as TineuVox~\citep{fang2022fast}, our approach integrates voxelized object-centric representations, which achieves an explicit understanding of the object geometries and physical interactions in the dynamic scenes and further facilitates downstream tasks such as scene editing.
}
\label{fig:intro}
\vspace{-5pt}
\end{figure*}

To address the first challenge and integrate object-centric learning into the volume rendering method, we propose a new form of object-centric representation that captures the local spatial structures of each object using time-aware voxel grids. This object-centric voxelization method explicitly allows us to \textit{infer a distribution of occupancy probabilities over objects at individual spatial locations}, thereby naturally facilitating 3D-consistent scene decomposition.
Unlike existing neural rendering techniques for dynamic scenes, such as TineuVox~\citep{fang2022fast}, our approach includes two key modifications in the model design (see Fig.~\ref{fig:intro}). 
First, it exploits the canonical-space deformation network to transform the voxelized object-centric representations over time. Second, based on the transformed occupancy probabilities per object, it employs a compositional NeRF for view-dependent image rendering, which also relies on a set of object-centric latent codes to represent the time-invariant features of each object.
Notably, in contrast to prior research that focuses on 3D dynamic scene reconstruction~\citep{fang2022fast, park2021hypernerf, Wu2022D2NeRFSD}, our unsupervised voxelization approach provides two additional advantages beyond novel view synthesis. First, it allows for fine-grained separation of object-centric information in 3D space, enabling a better understanding of the scene.
Second, it enables direct scene editing (\textit{e.g.}, object removal, replacement, and trajectory modification) that is not feasible in existing video decomposition methods. This is done by directly manipulating the voxel grids or the learned deformation function without the need for further training.

For the second challenge of semantics integration, we observe that extending our previous work of \model{}~\cite{zhao2024dynavol} to real-world data poses significant challenges due to the increased complexity of color and geometry patterns and the limited range of available view directions.
To address these issues, we present a new method that leverages 2D semantic feature maps extracted from an image encoding network trained on large-scale datasets. We develop a novel semantic volume slot attention module to 
\revise{project the learned 3D voxel grids onto 2D feature maps,}
This design enables our model to correlate the object-centric voxel grids with the pre-learned 2D feature maps, incorporating common knowledge of object geometries. This is particularly beneficial for handling previously unseen object structures, enhancing the model's ability to interpret complex real-world scenes.

In our experiments, we initially evaluate \newModel{} on simulated 3D dynamic scenes that contain different numbers of objects, diverse motions, shapes (such as cubes, spheres, and real-world shapes), and materials (such as rubber and metal). 
In the simulated dataset, we directly assess the performance of \newModel{} for scene decomposition by projecting the object-centric volumetric representations onto 2D planes and comparing it with existing approaches, such as SAVi~\citep{kipf2021conditional}, uORF~\citep{yu2021unsupervised}, and MovingParts~\cite{yang2024movingparts}.
Subsequently, we conduct a series of experiments on real-world scenes. Specifically, we compare \newModel{} against NeuralDiff~\citep{tschernezki2021neuraldiff}, HyperNeRF~\citep{park2021hypernerf}, TineuVox~\cite{fang2022fast}, and $\text{D}^2$NeRF~\citep{Wu2022D2NeRFSD} in novel view synthesis tasks, and against OCLR~\cite{xie2022segmenting} and ClipSeg~\cite{lueddecke22_cvpr} in scene decomposition tasks. These experiments validate the effectiveness of \newModel{}, highlighting the necessity of incorporating semantic features.

In summary, \newModel{} provides an early exploration of using NeRF-based models to tackle the research problem of unsupervised object-centric decomposition of 3D dynamic scenes. 
We extend our previous studies presented at ICLR 2024 in the following aspects:
\begin{itemize}[leftmargin=*]
\item \textit{Setup:} We adopt a more affordable experimental setting, requiring only training data captured by a monocular camera instead of using multiple viewpoints at the initial timestamp.
\item \textit{Method:} We introduce semantic features to enhance object-centric representation learning, significantly improving performance in real-world scenes.
\item \textit{Experiments:} We incorporate more real-world scenes and compare with more recent approaches, including TineuVox, MovingParts, OCLR, and ClipSeg, to demonstrate the superiority of \newModel{} in more challenging scenarios.
\end{itemize}

%% file: text/problem_setup.tex
\section{Problem Setup}



We assume a set of RGB images $\{\mathbf{I}_t, \mathbf{T}_t\}_{t=1}^T$ collected with a moving monocular camera, where $\mathbf{I}_t\in \mathbb{R}^{H\times W\times 3}$ are images acquired under camera poses $\mathbf{T}_t \in \mathbb{R}^{4\times 4}$, and $T$ is the length of the sequential video frames.
%
%
The goal is to decompose each object in the scene by harnessing the underlying space-time structures and semantics present in the visual data without additional supervision.

Please note that, unlike the previous work of DynaVol presented at ICLR 2024, we no longer use multiple viewpoints at the initial timestamp. 
We consider scenarios where (i) the training images are from viewpoints randomly sampled from the upper hemisphere, following the setup in D-NeRF~\cite{Pumarola2020DNeRFNR}, and (ii) a monocular camera captures the training images along a smooth moving trajectory, which is easily achievable in real-world applications.

%% file: text/method.tex
\section{Method}

In this section, we first review the original neural radiance field (NeRF) methodologies (Sec. \ref{sec:pre}).
Next, we introduce the concept of object-centric voxel representations and discuss the overall framework of our approach (Sec. \ref{sec:overview}). 
Subsequently, we present the details of each network component in the proposed \model{} and \newModel{} (Sec. \ref{sec:component}). 
Finally, we present the three-stage training procedure of our approach (Sec. \ref{sec:optimiz}).

\subsection{Preliminaries on NeRF}
\label{sec:pre}

The Neural radiance field models 3D scenes by mapping the 3D coordinates $\mathbf{x}$ and view directions $\mathbf{d}$ to the corresponding density $\sigma$ and color $\mathbf{c}$ using a neural network $\Phi$:
\begin{equation}
\begin{split}
(\sigma, \mathbf{c}) = \Phi(\mathbf{x}, \mathbf{d}).
\end{split}
\end{equation}
The outputs $\sigma$ and $\mathbf{c}$ are then transformed to the pixel values on a 2D image using volume rendering techniques~\citep{max1995optical}. 
Given a ray $\mathbf{r} = \mathbf{o} + t\mathbf{d}$ emitted from $\mathbf{x}$ to a specific pixel in the image, where $\mathbf{o}$ and $\mathbf{d}$ are respectively the origin and direction of the ray and $t$ is the distance from $\mathbf{x}$ to the camera. 
The color of the pixel is determined by the integral of $(\sigma_i,\mathbf{c}_i)$ of all the points along the ray within a predefined near bound $t_n$ and a far bound $t_f$, where $P$ denotes the number of sampled points, and $\delta_i$ denotes the distance between the $i$-th and $(i+1)$-th points:
\begin{equation}
\begin{split}
C(\mathbf{r}) =& \sum_{i=1}^P T_i\left(1-\exp(-\sigma_i\delta_i) \right)\mathbf{c}_i, \\
T_i =& \exp \left(- \sum_{j=1}^{i-1} \sigma_j \delta_j \right).
\end{split}
\label{eq:render}
\end{equation} 
%
In this work, we learn object-centric representations that evolve over time to drive a compositional NeRF renderer.

\begin{figure*}[t]
\begin{center}
\centerline{
\includegraphics[width=\linewidth]{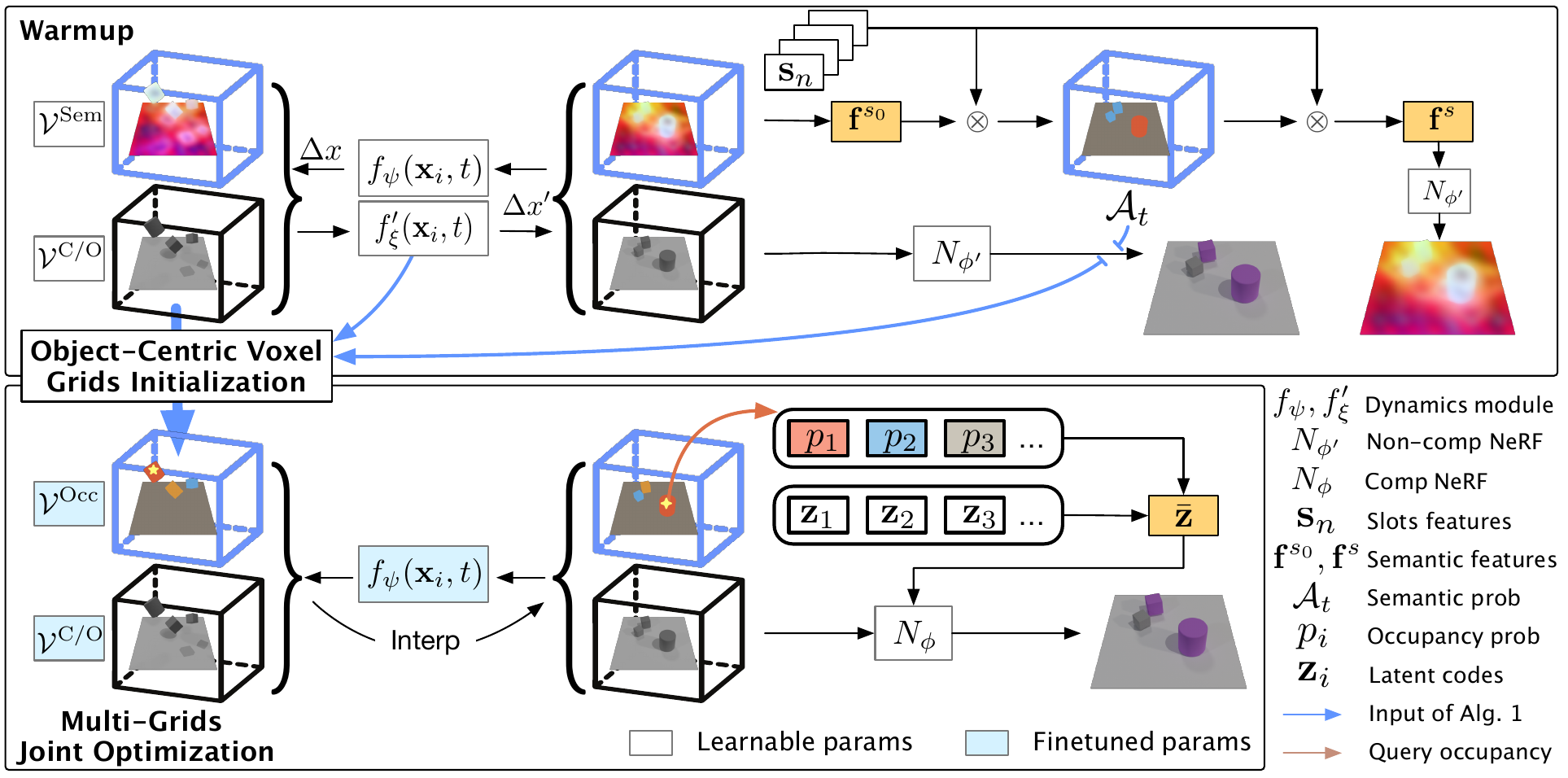}}
\vspace{-5pt}
\caption{
\newModel{} consists of three groups of network components: the bi-directional dynamics modules ($f_\psi$, $f_\xi^\prime$), the volume slot attention module, and neural rendering modules based on 3D and 4D voxels respectively. The training scheme involves a warmup stage (top), an object-centric voxel grids initialization stage whose pseudocode is given in Alg.~\ref{alg:initV}, and a multi-grids joint optimization stage (bottom). For clarity, \revise{we denote $\mathcal{V}^{\text{Color}}$ and $\mathcal{V}^{\text{Opac}}$ collectively as $\mathcal{V}^{\text{C/O}}$} and present the symbols and their descriptions in Table~\ref{tab:notation}.
}
\label{fig:arch}
\end{center}
\vspace{-10pt}
\end{figure*}

\begin{table*}[h!]
    \caption{Notations.}
    \vspace{-5pt}
    \label{tab:notation}
    \centering
    \begin{tabular}{>{\raggedright}p{2cm} p{6.5cm} >{\raggedright}p{2cm} p{5.8cm}}
    \toprule
    \textbf{Symbol} & \textbf{Description} & \textbf{Symbol} & \textbf{Description} \\
    \midrule
    \major{$\mathcal{V}^\text{Occ}$} & Voxel grids for occupancy probabilities per object & $\widehat{C}$ & Rendered 2D pixel colors \\
    \major{$\mathcal{V}^\text{Opac}$} & Voxel grids for opacity & $C$ & Observed 2D pixel colors \\
    \major{$\mathcal{V}^\text{Color}$} & Voxel grids for color-related features & $\mathcal{A}$ & Semantic probabilities \\
    \major{$\mathcal{V}^\text{Sem}$} & Voxel grids for semantic features & $\widehat{\mathcal{A}}$ & Rendered 2D semantic probabilities \\
    $\sigma$ & Opacity & $\widehat{F}$ & Rendered 2D semantic features \\
    $\mathbf{c}$ & Predicted color & $F$ & 2D semantic features extracted by DINOv2 \\
    $\mathbf{f}^c$ & Color-related features & $\{p_n\}_{n=1}^N$ & Occupancy probabilities per object \\
    $\mathbf{f}^s$ & Semantic features & $\widehat{P}$ & Rendered 2D occupancy probabilities \\
    $N$ & Number of latent codes 
    &$\bar{P}$ & Pseudo-labels obtained by CRF   \\
    $\{\mathbf{z}_n\}_{n=1}^N$ & Object-centric latent codes &
    $f_\psi, f_\xi^\prime$ & Bi-directional deformation networks  \\
    $\mathbf{S}$ / $\{\mathbf{s}_n\}_{n=1}^N$ & Object-centric slot features &
    $\mathcal N_\phi, \mathcal N_{\phi^\prime}$ & Compositional/Non-compositional NeRF\\
    \bottomrule
    \end{tabular}
\end{table*}

\subsection{Object-Centric Voxelization}
\label{sec:overview}

\newModel{} is trained in an inverse graphics framework to synthesize $\{\mathbf{I}_t\}_{t=1}^T$ without any further supervision.
%
Formally, the goal is to learn an object-centric projection of $(\mathbf{x},\mathbf{d},t, \{\mathbf{z}_n\}_{n=1}^N, \major{\mathcal{V}^\text{Occ}}, \major{\mathcal{V}^\text{Opac}}, \major{\mathcal{V}^\text{Color}})\xrightarrow[]{}(\mathbf{c},\sigma)$. 
$\mathbf{x}=(x,y,z)$ is a 3D point, $\{\mathbf{z}_n\}_{n=1}^N \in \mathbb{R}^{N \times D_z}$ denotes a set of learnable latent codes associated with individual objects, and $N$ is the predefined number of the latent codes that is assumed to be larger than the number of objects.
$\major{\mathcal{V}^\text{Occ}}, \major{\mathcal{V}^\text{Opac}}, \major{\mathcal{V}^\text{Color}}$ are voxelized features, which respectively represent occupancy probabilities per object, opacity, and color-related features.

To understand the scene from an object-centric perspective, we extend the classic approach of using 3D voxel grids for volume rendering by incorporating 4D voxel grids, represented as $\major{\mathcal{V}^\text{Occ}} \in \mathbb{R}^{N \times N_x\times N_y\times N_z}$, where $(N_x, N_y, N_z)$ denote the spatial resolutions.
%
%
The newly added dimension in the expanded voxel features captures the occupancy probabilities for each object at the grid vertices. 
We have 
\begin{equation}
\mathrm{Interp}(\mathbf{x}, \major{\mathcal{V}^\text{Occ}}): \ (\mathbb{R}^3, \mathbb{R}^{N \times N_x\times N_y\times N_z}) \xrightarrow{} \mathbb{R}^{N},
\end{equation}
which means the occupancy probabilities $\{p_n\}_{n=1}^N$ at any given 3D position $\mathbf{x}$ within the voxel grid are efficiently retrieved using trilinear interpolation.
Following this, we apply the Softmax activation function to the interpolated values to normalize $\{p_n\}_{n=1}^N$, ensuring they sum to one and are proportionately scaled.

In addition to $\major{\mathcal{V}^\text{Occ}}$, our approach employs three other voxelized representations that are independent of the object index: $\major{\mathcal{V}^\text{Opac}} \in \mathbb{R}^{1\times N_x\times N_y\times N_z}$ for modeling the opacity $\sigma$, $\major{\mathcal{V}^\text{Color}} \in \mathbb{R}^{D_c \times N_x\times N_y\times N_z}$ for modeling color-related features $\mathbf{f}^c$, and $\major{\mathcal{V}^\text{Sem}} \in \mathbb{R}^{D_s \times N_x\times N_y\times N_z}$ for modeling semantic features of the scene.
The incorporation of semantic features is a pivotal contribution of the improved \newModel{}.
%
We optimize the initial states of $\{\major{\mathcal{V}^\text{Opac}}, \major{\mathcal{V}^\text{Color}}, \major{\mathcal{V}^\text{Sem}}, \major{\mathcal{V}^\text{Occ}}\}$ and learn a deformation module $f_\psi(\cdot)$ to transform them over time in the canonical space.

%
%

Apart from the voxelized representations that present the local geometry and appearance of the scene, our approach also learns a set of object-centric latent codes $\{\mathbf{z}_n\}_{n=1}^N$ to capture the global object-specific information and particularly use them in the NeRF renderer.
We set $\{\mathbf{z}_n\}_{n=1}^N$ as time-invariant features.



\subsection{Model Components}
\label{sec:component}

As shown in Fig.~\ref{fig:arch}, \newModel{} consists of three modules:
\begin{enumerate}[label=(\roman*)]
    \item The bi-directional deformation networks, $f_\psi$ and $f_\xi^\prime$, which learn the canonical-space transitions over time.
    \item The semantic volume slot attention module, 
    which learn the semantic features $\major{\mathcal{V}^\text{Sem}}$ and disentangle them into object-centric slot features $\mathbf{S}$ respectively. We extract a set of 2D semantic-level features $\{\mathbf{F}_t\in \mathbb{R}^{C \times H^\prime \times W^\prime}\}_{t=1}^T$ from $\{\mathbf{I}_t\}_{t=1}^T$
    using a pre-trained 2D feature extractor, \textit{e.g.}, DINOv2~\cite{oquab2023dinov2} as additional learning objectives for this module.
    \item The neural renders $\mathcal N_\phi$ and $\mathcal N_{\phi^\prime}$. $\mathcal N_\phi$ is a compositional NeRF that jointly uses $\major{\mathcal{V}^\text{Opac}}, \major{\mathcal{V}^\text{Color}}, \major{\mathcal{V}^\text{Occ}}$, and $\{\mathbf{z}_n\}_{n=1}^N$ for rendering of each object. $\mathcal N_{\phi^\prime}$ is a non-compositional NeRF only used to initialize the voxel representations.
\end{enumerate}

\subsubsection{Bi-Directional Deformation Networks} 
\myparagraph{Canonical-space dynamics modeling.} 
We leverage the canonical space for dynamics modeling, where the canonical coordinates are used in the Hamiltonian formulation of classical mechanics to describe a physical system at any given point in time.
Specifically, as illustrated in Fig.~\ref{fig:arch}, we use a dynamics module $f_\psi(\mathbf{x}_i, t)$ to learn the deformation field, where $\mathbf{x}_i$ and $t$ are first projected into higher dimensions through positional encoding.
This module transforms a 3D point $\mathbf{x}_i$ at an arbitrary time to its corresponding position at the initial time by predicting a position movement $\Delta \mathbf{x}_i$ and calculating $\mathbf{x}_i^\prime=\mathbf{x}_i+\Delta \mathbf{x}_i$.
Using canonical-space mechanics, we determine the variations in voxel representations over time.
For any given coordinate $(\mathbf{x}_i,t)$, we have:
\begin{equation}
\small
\begin{split}
\sigma_i = \mathrm{Interp}&\left(\mathbf{x}_i^\prime, \major{\mathcal{V}^\text{Opac}}\right), \\
\mathbf{f}_i^c=\mathrm{Interp}\left(\mathbf{x}_i^\prime, \major{\mathcal{V}^\text{Color}}\right)& \oplus \ldots \oplus \mathrm{Interp}\left(\mathbf{x}_i^\prime, \major{\mathcal{V}^\text{Color}}\left[:: s_M\right]\right).
\end{split}
\label{eq: color-feature}
\end{equation}
The opacity is retrieved through trilinear interpolation. For the color-related voxel features, we adopt a multi-distance interpolation strategy \cite{fang2022fast}, where $\oplus$ denotes concatenating features across various resolutions and $[::s_M]$ indicates the stride used to sample the grid at various levels. We set $s_M = \{2,4\}$ in our experiments.
The retrieved voxel features are then used in the neural renderer.

\myparagraph{Forward dynamics modeling.} 
Additionally, we employ another dynamics module $f_\xi^\prime(\mathbf{x}_i, t)$ to capture the forward movement $\Delta x_i^\prime$ from the initial time to timestamp $t$. 
%
\revise{This module enables the calculation of a cycle-consistency loss, as defined in Eq. \eqref{eq:cyc-loss}, which has been shown in previous literature~\cite{Liu2022DeVRFFD, yang2024movingparts} to improve the coherence of learned canonical-space transitions. We further discuss the validity of the cycle-consistency training approach in Sec. \ref{sec:warm}.}
As a result, $f_\xi^\prime(\mathbf{x}_i, t)$ provides useful information, \revise{\textit{i.e.}, the positions and velocities of each voxel at time $t$}, for the subsequent connect components algorithm in the warmup stage (detailed in Sec. \ref{sec:warm}), which can improve the initialization of the voxel grids.






\subsubsection{Semantic Volume Slot Attention}
To extract semantic features and leverage them to improve scene decomposition by correlating them with voxel grids, we design a simple yet effective module named \textit{semantic volume slot attention}.

\myparagraph{Semantic feature extractor.}
We use DINOv2~\cite{oquab2023dinov2} as the semantic feature extractor and further employ FeatUp~\cite{fu2024featup} to generate higher-resolution feature maps.
Notably, both methods are trained in an unsupervised manner.
We particularly use the semantic features in real-world scenes, while our previous work on \model{} does not exploit any semantic features.
For synthetic scenes, instead, we observe that our approach can effectively handle the object geometries even in the absence of semantic features.


\myparagraph{Volume slot attention.}
During the warmup phase, we employ the volume slot attention to progressively derive a set of global time-invariant object-level representations from the local, time-varying voxel features. 
Specifically, we introduce the semantic voxel grids, $\major{\mathcal{V}^\text{Sem}} \in \mathbb{R}^{D_s \times N_x \times N_y \times N_z}$, and a set object-centric ``slots features'', $\mathbf{S}\in \mathbb{R}^{N\times D_s}$, borrowed from the prior literature on 2D static scene decomposition~\citep{locatello2020object}. 
Given a 3D point $\mathbf{x}$ at timestamp $t$, we have:
\begin{equation}
\begin{split}
    \major{\mathbf{f}_i^{s_0}} \in \mathbb{R}^{D_s} &= \text{Interp}\left(\mathbf{x}_i+f_\psi(\mathbf{x}_i, t),\major{\mathcal{V}^\text{Sem}}\right) \\
    \mathcal{A}_i \in \mathbb{R}^{N} &= \mathrm{softmax}_{N} \left(\frac{\major{\mathbf{f}_i^{s_0}} \cdot \mathbf{S}^T}{\tau}\right) \\
    \mathbf{f}_i^s \in \mathbb{R}^{D_s} &= \mathcal{A}_i \cdot \mathbf{S},
\label{eq:semantic_attn}
\end{split}
\end{equation}
%
where $T$ denotes the transpose of a matrix, $\tau$ is a learnable temperature factor, and $\mathcal{A}_i$ represents the semantic probabilities that are further used to initialize the 4D voxel grids, $\major{\mathcal{V}^\text{Occ}}$. Unlike prior slot attention methodologies, we do not use query, key, and value projections. This approach improves efficiency and ensures that linear projections do not affect the correlation between features and slots. 
Subsequently, $\mathbf{f}^s_i$ and $\mathcal{A}_i$ are used to obtain the 2D semantic features and corresponding 2D maps of semantic probabilities with the quadrature rule~\citep{max1995optical}:
\begin{equation}
\begin{split}
    \widehat{F}(\mathbf{r}) &= \sum_{i=1}^P T_i\left(1-\exp(-\sigma_i\delta_i)\right)\mathbf{f}_i^s \\ 
    \widehat{\mathcal{A}}(\mathbf{r}) &= \sum_{i=1}^P T_i\left(1-\exp(-\sigma_i\delta_i)\right)\mathcal{A}_i.
\label{eq:sem_feature}
\end{split}
\end{equation}
We then use a linear layer to map $\widehat{F}(\mathbf{r})$ from $\mathbb{R}^{D_s}$ to $\mathbb{R}^C$, the channel dimension of the 2D semantic feature maps provided by DINOv2, and compute the reconstruction loss as described in Eq.~\eqref{eq:semantic_loss}.
Additionally, we supervise the accumulated $\widehat{\mathcal{A}}(\mathbf{r})$ using the entropy loss, as detailed in Eq.~\eqref{eq:semantic_loss}, to further reduce the ambiguities of the learned semantic representations. 
%
%
For an empirical analysis of such model designs, please refer to Sec.~\ref{analysis}.

\subsubsection{Object-Centric Renderer}

Previous compositional NeRF models, like uORF~\citep{yu2021unsupervised}, typically use MLPs to learn a continuous mapping from sampling point, viewing direction, and slot features to the emitted densities and colors of different objects.
%
In contrast, our neural renderer takes as inputs $\{\mathbf{z}_n\}_{n=1}^N$, which is a set of learnable parameters representing the global object-centric features.
%
%
%
In \newModel{}, we learn the linear combinations of $\{\mathbf{z}_n\}_{n=1}^N$ to construct the inputs of the rendering network, such that:
\begin{equation}
\begin{split}
    \major{p_i =}& \ \major{\mathrm{softmax}_N(\mathrm{Interp}(\mathbf{x}_i+f_\psi(\mathbf{x}_i, t), \major{\mathcal{V}^\text{Occ}}),} \\ 
    \overline{\mathbf{z}}_i =& \sum_{n=1}^N p_i[n] \cdot \mathbf{z}_n.
\end{split}
\end{equation}
Subsequently, we use an MLP to learn object-centric projections $\mathcal N_\phi: (\mathbf{f}_i^c, \mathbf{d}, \overline{\mathbf{z}}_i)\xrightarrow{}\mathbf{c}_i$, where $\mathbf{f}_i^c$ are color-related features from Eq.~\eqref{eq: color-feature}.
We apply positional encodings to $\mathbf{f}_i^c$ and $\mathbf{d}$ before feeding them to the MLP.
In the rendering process, we directly query the opacity $\sigma$ from the voxel grids $\major{\mathcal{V}^\text{Opac}}$ within the canonical space. 
We follow Eq.~\eqref{eq:render} to estimate the pixel color of a sampling ray using the predicted $\mathbf{c}_i$ and the retrieved $\sigma_i$.

\revise{In addition to the object-centric renderer, we also introduce a non-compositional NeRF, denoted by $\mathcal N_{\phi^\prime}$. 
Unlike $\mathcal N_\phi$, which is used in the multi-grids joint optimization stage, $\mathcal N_{\phi^\prime}$ is used in the warmup stage to initialize the voxel representations.
Specifically, it employs an MLP to learn a mapping function: 
$(\mathbf{f}_i^c, \mathbf{d})\xrightarrow{}\mathbf{c}_i$, which is unrelated to any object-centric features. 
}

\subsection{Training Procedure}
\label{sec:optimiz}

\newModel{} involves three stages in the training phase:
First, a warmup stage that learns to obtain a basic understanding of the geometric, semantic, and dynamic priors.
Second, an initialization stage for the voxelized occupancy probabilities $\major{\mathcal{V}^\text{Occ}}$, involving the connected components algorithm.
Third, an end-to-end optimization stage that jointly optimizes $\{\major{\mathcal{V}^\text{Opac}}, \major{\mathcal{V}^\text{Color}}, \major{\mathcal{V}^\text{Occ}}, \{\mathbf{z}_n\}_{n=1}^N, f_\psi\}$ to refine the voxelized object occupancy.


\begin{algorithm}[t]
\small
 \setstretch{1.2}
    \caption{Object-Centric Voxel Grids Initialization}
    \label{alg:initV}
    \begin{algorithmic}[1]
        \STATE \textbf{Input:} $\{\sigma_k\}_{k=1}^{N_s}$, $\{\mathbf{x}_k\}_{k=1}^{N_s}$, $\{\mathbf{c}^{\text{ind}}_k\}_{k=1}^{N_s}$, $\{y_k\}_{k=1}^{N_s}$, hyperparameters $\delta_{\text{den}}$, $\delta_{\text{pos}}$,
        $\delta_{\text{rgb}}$, $\delta_{\text{vel}}$, $N$ 
        \STATE \textbf{Output:} $\mathcal{V}^{\text{M}}$
        \STATE $\mathcal{N} = \{k \mid \sigma_k > \delta_{\text{den}},k \in [1, N_s] \}$ \major{\textit{\COMMENT{Filter out invalid locations}}}
        \FOR{$t\xleftarrow{}1$ \textbf{to} $T-1$}
                \STATE Calculate velocity $ \mathbf{v}_{t,j} = f^\prime_\xi(\mathbf{x}_j, t+1)-f^\prime_\xi(\mathbf{x}_j, t)$ \textbf{for} $j\in \mathcal{N}$ \major{\textit{\COMMENT{Get velocity information}}}
        \ENDFOR

        \FOR{Neighbor$(p, q)\in\{(u, v)\mid u,v \in \mathcal{N}, l_2(\mathbf{x}_u,\mathbf{x}_v)<\delta_\text{pos}\}$}
            \STATE $D_{\text{rgb}}(p,q) = l_2(\mathbf{c}^\text{ind}_p, \mathbf{c}^\text{ind}_q)$ 
            \STATE$D_{\text{vel}}(p,q) = \text{max}(l_2(\mathbf{v}_{t,p},  \mathbf{v}_{t,q})$ \textbf{for} $t\xleftarrow{}1$ \textbf{to} $T-1)$ 
            \IF{$D_{\text{rgb}}(p,q)<\delta_{\text{rgb}}$ and $D_{\text{vel}}(p,q)<\delta_{\text{vel}}$  and $y_p = y_q$ }
            \STATE $E(p, q) = 1 $   \major{\textit{\COMMENT{Add $E(p, q)$ to the edge set $\mathcal E$}}}
            \ELSE
            \STATE $E(p, q) = 0$
            \ENDIF
        \ENDFOR
        \STATE Generate the feature graph $G=(\mathcal{N},\mathcal E)$ 
        \STATE Obtain the cluster set $\{\text{cl}_n\}_{n=1}^{M} = \text{ConnectedComponents}(G)$
        \STATE $\{\text{cl}_n\}_{n=1}^{N} = \text{sort-and-interp}(\{\text{cl}_n\}_{n=1}^{M})$ 
        \major{\textit{\COMMENT{Sort clusters and interpolate the rest}}}
        \FOR{$n\xleftarrow{}1$ \textbf{to} $N$} 
            \STATE $\major{\mathcal{V}^\text{Occ}}[n,:,:,:] = \text{one-hot}(\text{cl}_k)$ \major{\textit{\COMMENT{Generate 4D voxel grids}}}
        \ENDFOR 
        \STATE Rendering 2D masks $\widehat{P}(\mathbf{r})$ with $\major{\mathcal{V}^\text{Occ}}$  
        \STATE Obtaining pseudo-labels by CRF: $\bar{P}(\mathbf{\mathbf{r}}) = \text{CRF}(\widehat{P}(\mathbf{r}), C(\mathbf{r}))$ 
    \end{algorithmic}
\end{algorithm}

\subsubsection{Warmup}
\label{sec:warm}
To reduce the difficulty of learning the object-centric 4D voxel grids $\major{\mathcal{V}^\text{Occ}}$, we first use $\{\mathbf{I}_t\}_{t=1}^T$ and $\{\mathbf{F}_t\}_{t=1}^T$ to warm-up the density grids $\major{\mathcal{V}^\text{Opac}}$, the color-related feature grids $\major{\mathcal{V}^\text{Color}}$, the semantic feature grids $\major{\mathcal{V}^\text{Sem}}$, the slot features $\mathbf{S}$, and the dynamics module $f_\psi(\cdot)$.
Additionally, we train an extra forward dynamics module $f^\prime_\xi(\cdot)$ and a non-compositional neural renderer $N_{\phi^\prime}(\cdot)$.
%
%
The overall objective is defined as $\mathcal{L}_\text{Warm} = \mathcal{L}_\text{Render} + \alpha_p \mathcal{L}_\text{Point} + \alpha_b \mathcal{L}_\text{Bg-Ent} + \alpha_c \mathcal{L}_\text{Cyc} + \mathcal{L}_\text{Recon} + \alpha_e \mathcal{L}_\text{Ent}$. 
This framework divides its focus between scene reconstruction, covered by the first four terms, and the representation of semantic features, addressed by the last two terms. The hyperparameter values of scene reconstruction are adopted from prior literature~\cite{Liu2022DeVRFFD}.
%
At a specific time, we define
\begin{equation}
\begin{split}
    &\mathcal{L}_\text{Render} = \frac{1}{|\mathcal{R}|}\sum_{{r}\in\mathcal{R}}\left\|\widehat{C}(\mathbf{r})-C(\mathbf{r})\right\|_2^2 
    \\
    &\mathcal{L}_\text{Bg-Ent} = \frac{1}{|\mathcal{R}|}\sum_{{r}\in\mathcal{R}}-\widehat{w}_l^r  \log(\widehat{w}_{l}^r) - (1-\widehat{w}_{l}^r) \log(1-\widehat{w}_{l}^r)\\
    &\mathcal{L}_\text{Point} = \frac{1}{|\mathcal{R}|}\sum_{{r}\in\mathcal{R}}\left(\frac{1}{P}\sum_{i=1}^{P}\left\|\overline{\mathbf{c}}_i - C(\mathbf{r})\right\|_2^2\right),   \\
\end{split}
\label{eq:loss}
\end{equation}
where $\mathcal{L}_\text{Render}$ is the rendering loss $\mathcal{L}_\text{Render}$ between the predicted and observed pixel colors, $\mathcal{L}_\text{Bg-Ent}$ is the background entropy loss that encourages the renderer to concentrate on either foreground or background, and $\mathcal{L}_\text{Point}$ is the per-point RGB loss following DVGO~\citep{sun2022direct}. 
$\mathcal{R}$ is the set of sampled rays in a batch, $P$ is the number of sampling points along ray $r$, and $\widehat{w}_l^r$ is the color contribution of the last sampling point obtained by $\widehat{w}^r_l = T_{P}(1-\exp(-\sigma_{P} \delta_{P}))$.
We use the above loss functions as basic objective terms for neural rendering.

To enhance the coherence of the learned canonical-space dynamics, we train an additional module $f^\prime_\xi(\cdot)$ using the following cycle-consistency loss:
\begin{equation}
    \ \mathcal{L}_\text{Cyc} = \frac{1}{|\mathcal{R}|}\sum_{{r}\in\mathcal{R}}\left(\frac{1}{P}\sum_{i=1}^{P}\left\| f_{\psi} (x_i,t) + f^\prime_{\xi} (x^\prime_i,t)\right\|_2^2\right),
\label{eq:cyc-loss}
\end{equation}
where $x^\prime_i=x_i+f_{\psi} (x_i,t), i\in[1, P]$.
\revise{Notably, since both networks, $f^\prime_\xi(\cdot)$ and $f_\psi(\cdot)$, are jointly trained during the same stage, it raises the question of whether they can be correctly trained together. In Sec.~\ref{analysis}, we provide further discussions and experimental analyses to address this concern.}


\begin{table*}[t]
\caption{Novel view synthesis results of our approach compared with D-NeRF, DeVRF, TineuVox, and a variant of DeVRF aligned with our experimental configurations (see text for details). We show results averaged over $60$ novel views (one view per timestamp).
}
\label{tab:nv_ssim}
\vspace{-10pt}
\begin{center}
\setlength{\tabcolsep}{14pt}{}
\begin{small}
\begin{tabular}{lcccccccc}
\toprule
& \multicolumn{2}{c}{3ObjFall} &\multicolumn{2}{c}{3ObjRand} &\multicolumn{2}{c}{3ObjMetal} &\multicolumn{2}{c}{3Fall+3Still} \\
Method
& PSNR$\uparrow$  & SSIM$\uparrow$ 
& PSNR$\uparrow$  & SSIM$\uparrow$ 
& PSNR$\uparrow$  & SSIM$\uparrow$ 
& PSNR$\uparrow$  & SSIM$\uparrow$ \\
\cmidrule(lr){1-1}  \cmidrule(lr){2-3}  \cmidrule(lr){4-5}  \cmidrule(lr){6-7}  \cmidrule(lr){8-9}
D-NeRF        &28.54&0.946    &12.62&0.853    &27.83&0.945    &24.56&0.908\\
DeVRF          &24.92&0.927    &22.27&0.912    &25.24&0.931    &24.80&0.931\\
DeVRF-Dyn           &18.81&0.799    &18.43&0.799    &17.24&0.769    &17.78&0.765\\
TineuVox  &\textbf{32.79}&\textbf{0.968}  &29.46&0.955  &\underline{30.84}&\underline{0.960} &\underline{29.61} & \underline{0.950}\\
DynaVol &31.83&\underline{0.967}   &\underline{31.10}&\underline{0.966} 
&29.28&0.954   &27.83&0.942\\ 
DynaVol-S
&\underline{32.15} &0.963    &\textbf{32.58}&\textbf{0.973}   &\textbf{30.87}&\textbf{0.963}    &\textbf{31.59}&\textbf{0.966}\\
\midrule
& \multicolumn{2}{c}{6ObjFall} &\multicolumn{2}{c}{8ObjFall} &\multicolumn{2}{c}{3ObjRealSimp} &\multicolumn{2}{c}{3ObjRealCmpx} \\
Method
& PSNR$\uparrow$  & SSIM$\uparrow$ 
& PSNR$\uparrow$  & SSIM$\uparrow$ 
& PSNR$\uparrow$  & SSIM$\uparrow$ 
& PSNR$\uparrow$  & SSIM$\uparrow$ \\
\cmidrule(lr){1-1}  \cmidrule(lr){2-3}  \cmidrule(lr){4-5}  \cmidrule(lr){6-7} \cmidrule(lr){8-9}
D-NeRF           &28.27&0.940    &27.44&0.923    &27.04&0.927   &20.73&0.864\\
DeVRF          &24.83&0.905    &24.87&0.915    &24.81&0.922    &21.77&0.891\\
DeVRF-Dyn           &17.35&0.738    &16.19&0.711    &18.64&0.717    &17.40&0.778\\
TineuVox  &\underline{30.62}&\underline{0.955}  &\underline{31.27}&\underline{0.952}  &\underline{30.70}&\underline{0.954} &\underline{27.72} & \underline{0.927}\\
DynaVol  &29.70&0.948  &29.86&0.945   &30.20&0.952  &26.80&0.917\\ 
DynaVol-S        &\textbf{31.37}&\textbf{0.961}    &\textbf{31.78}&\textbf{0.961}    &\textbf{30.93}&\textbf{0.958}    &\textbf{28.82}&\textbf{0.941}\\
\bottomrule
\end{tabular}
\end{small}
\end{center}
\vspace{-5pt}
\end{table*}

Furthermore, we incorporate the reconstruction loss $\mathcal{L}_{\text{Recon}}$ between the predicted and observed pixel semantic features, as well as the entropy loss $\mathcal{L}_{\text{Ent}}$ on the semantic probabilities to regularize them towards a one-hot distribution:
\begin{equation}
\begin{split}
    &\mathcal{L}_\text{Recon} = \frac{1}{|\mathcal{R}|}\sum_{{r}\in\mathcal{R}}\left\|\widehat{F}(\mathbf{r})-F(\mathbf{r})\right\|_1 
    \\
    &\mathcal{L}_\text{Ent} = \frac{1}{|\mathcal{R}|}\sum_{{r}\in\mathcal{R}} \sum_{n=1}^N -\widehat{\mathcal{A}}(\mathbf{r})_n \log(\widehat{\mathcal{A}}(\mathbf{r})_n).\\
\end{split}
\label{eq:semantic_loss}
\end{equation}

\subsubsection{Object-Centric Voxel Grids Initialization}

We initialize the 4D object-centric grids $\mathcal{V}^{\text{M}}$ with two steps: (1) the feature graph generation and connected components computation and (2) a post-processing step with Conditional Random Field (CRF).
As illustrated in Alg. \ref{alg:initV}, the entire process takes as inputs the voxel position set $\{\mathbf{x}_k\}$ with a size of $N_s=N_x \times N_y \times N_z$, the voxel density set $\{\sigma_k\}$ from $\major{\mathcal{V}^\text{Opac}}$, the view-independent colors $\{\mathbf{c}_k^{\text{ind}}\}$ generated by $N_{\phi^\prime}$, and the corresponding semantic labels $\{y_k\}$ obtained by $\mathrm{argmax}_N\left(\mathcal{A}\right)$ in Eq.~\eqref{eq:semantic_attn}.
We also consider the forward canonical-space transitions generated by $f^\prime_\xi(\cdot)$.
Besides, we set five hyperparameters in advance, where $\delta_{\text{den}}$ represents the density value threshold, $\delta_{\text{pos}}$ represents the position distance threshold, $\delta_{\text{rgb}}$ represents the RGB distance threshold, $\delta_{\text{vel}}$ represents the velocity distance threshold, and $N$ represents the maximum number of objects.

Specifically, we first filter out invalid locations from $\{\mathbf{x}_k\}$ with density values below the predefined threshold $\delta_{\text{den}}$. The remaining voxels are set as the nodes of a feature graph $G$. 
Next, we create the edges between adjacent graph nodes in $G$ based on their similarities in position, color, velocity, and semantics, where each node retains features of $(\mathbf{x}_k, \mathbf{c}_k^{\text{ind}}, \mathbf{v}_{1:T-1,k}, y_k)$ of the valid voxels.
%
%
Subsequently, we perform the connected components algorithm on $G$, where the $M$ output clusters symbolize different objects, based on our assumption that voxels of the same object share similar motion, appearance, and semantics. 
If $M \leq N$, we directly utilize the clustering results and keep the remaining $N-M$ clusters empty. If $M > N$, we prioritize the clusters based on their contributions to the rendering results, as described in~\cite{li2023compressing}. We initialize $\mathcal{V}^M$ by retaining the top $N$ clusters and applying nearest neighbor interpolation to the remaining voxel grids.

Furthermore, as an optional post-processing step in \newModel{}, we employ a Conditional Random Field (CRF) to refine $\mathcal{V}^M$, following the common practice in video object segmentation, this step is beneficial for mitigating potential noise introduced by the connected components algorithm, thereby smoothing the segmentation results. The CRF method leverages RGB information to refine 2D mask predictions. Although this step could potentially compromise the 4D consistency of segmentation results, we address this by using the refined masks as pseudo-labels in the subsequent stage. This allows us to finetune the 4D voxel grids, $\major{\mathcal{V}^\text{Occ}}$, thereby ensuring consistency in object-centric representations.

\subsubsection{Multi-Grids Joint Optimization}
In this stage, we finetune the dynamics module $f_\psi$, the geometric grids $\major{\mathcal{V}^\text{Opac}}$, the appearance grids $\major{\mathcal{V}^\text{Color}}$, and the object-centric grids $\major{\mathcal{V}^\text{Occ}}$ obtained in the first two stages. 
%
%
We finetune the parameters of $(\major{\mathcal{V}^\text{Opac}}, \major{\mathcal{V}^\text{Color}}, \major{\mathcal{V}^\text{Occ}}, \psi)$ and begin training $(\{\mathbf{z}_n\}_{n=1}^N, \phi)$ from scratch.
Following an end-to-end training scheme, the dynamics module, object-centric latent codes, and compositional renderer collaboratively contribute to object-centric dynamic scene understanding.
%
%
%
%
%
%
The loss function in this stage is defined as
    $\mathcal{L}_\text{Dyn} =  \mathcal{L}_\text{Render} + \alpha_p \mathcal{L}_\text{Point} + \alpha_b \mathcal{L}_\text{Bg-Ent}$,
complemented by a cross-entropy loss between the rendered 2D segmentations $\widehat{P}(\mathbf{r})$ from $\major{\mathcal{V}^\text{Occ}}$ and the pseudo-labels $\bar{P}(\mathbf{r})$ derived from the second stage: $\mathcal{L}_\text{CE} = -\sum_{n=1}^N \bar{P}(\mathbf{r})_n \log \widehat{P}(\mathbf{r})_n $. 


%% file: text/expri.tex
\section{Experiments}


\begin{table*}[t]
\caption{Novel view synthesis results using real-world scenes from HyperNeRF and $\text{D}^2$NeRF. On average (as shown in the last column), our approach outperforms HyperNeRF by $10.8\%$ in PSNR and $8.3\%$ in SSIM. Additionally, \newModel{} demonstrates better performance compared to $\text{D}^2$NeRF, another neural renderer that incorporates disentanglement learning.
}
\label{tab:real_world_synthesis}
\vspace{-10pt}
\begin{center}
\begin{small}   
\setlength{\tabcolsep}{6pt}{}
\begin{tabular}{lcccccccccccc}
\toprule
& \multicolumn{2}{c}{Chicken} &\multicolumn{2}{c}{Broom} &\multicolumn{2}{c}{Peel-Banana} &\multicolumn{2}{c}{Duck} 
&\multicolumn{2}{c}{Balloon}
&\multicolumn{2}{c}{Avg.}\\
Method
& PSNR$\uparrow$   & SSIM$\uparrow$
& PSNR$\uparrow$   & SSIM$\uparrow$
& PSNR$\uparrow$   & SSIM$\uparrow$
& PSNR$\uparrow$   & SSIM$\uparrow$
& PSNR$\uparrow$   & SSIM$\uparrow$
& PSNR$\uparrow$   & SSIM$\uparrow$\\
\cmidrule(lr){1-1}  \cmidrule(lr){2-3}  \cmidrule(lr){4-5}  \cmidrule(lr){6-7} 
\cmidrule(lr){8-9} \cmidrule(lr){10-11} \cmidrule(lr){12-13}
NeurDiff      &21.17&0.822    &17.75&0.468    &19.43&0.748 &21.92&0.862 &20.13&0.836 &20.08 &0.747 \\
HyperNeRF &26.90&\underline{0.948}    &19.30&0.591   &22.10&0.780       &20.64&0.830  &17.81&0.803   &21.35&0.790 \\
$\text{D}^2$NeRF            &24.27&0.890    &\underline{20.66}&\textbf{0.712}  &21.35&0.820     &\underline{22.07}&0.856  &20.92&0.858 &21.85 &0.827\\
TineuVox &\underline{28.30} &0.947 &21.43&0.686  &\underline{24.40}&\underline{0.873} &21.85 &\underline{0.895}  &21.04&0.850 &\underline{23.40}&\underline{0.850}\\
DynaVol         &27.01&0.934  &\textbf{21.49}&\underline{0.702}    &24.07&0.863       &21.43&0.878  &\underline{21.07}&\underline{0.863}  &23.01&0.848 \\
DynaVol-S &\textbf{28.43}&\textbf{0.949} &20.98&0.671 &\textbf{25.11}&\textbf{0.894} &\textbf{22.40}&\textbf{0.902} &\textbf{21.38}&\textbf{0.864} &\textbf{23.66}&\textbf{0.856}\\
\bottomrule
\end{tabular}
\end{small}
\end{center}
\vspace{-10pt}
\end{table*}

\begin{figure*}[t]
    \centering
    \includegraphics[width=0.99\textwidth]{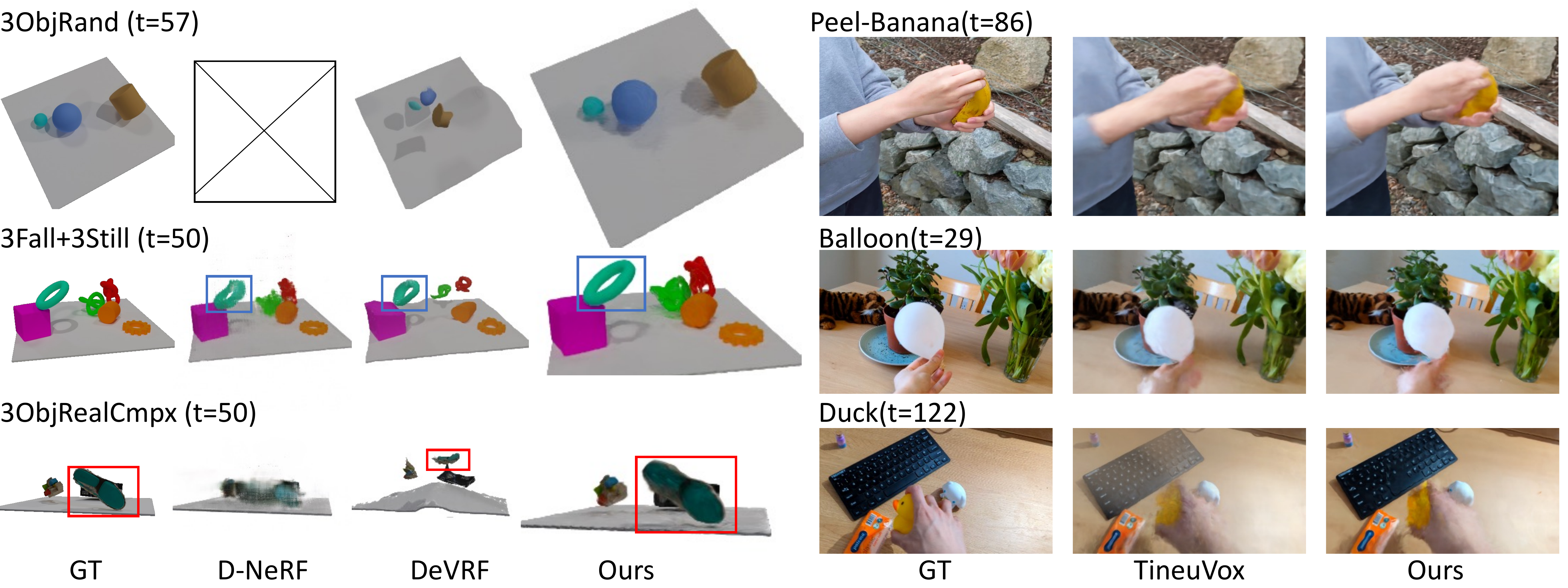}
    \vspace{-5pt}
    \caption{Novel view synthesis results. D-NeRF fails in the left synthetic examples on 3ObjRand. In the right real-world scenes, \newModel{} significantly outperforms the prior art, TineuVox, by learning object-centric features.}
\label{fig:novel_view}
\vspace{-5pt}
\end{figure*}

\subsection{Experimental Setup}

\myparagraph{Implementation details.} 
We set the size of voxel grids to $160^3$, the assumed number of maximum objects to $N=15$ for real-world scenes and $10$ for synthetic scenes, the dimension of semantic features to $D_s=16$, the dimension of latent codes to $D_z = 64$, and the dimension of color-related features to $D_c = 6$.
We use $2$ hidden layers with $256$ channels in the renderer,
and use the Adam optimizer with a batch of $4{,}096$ rays in the two training stages. 
The base learning rates are $0.1$ for the voxel grids and $1e^{-3}$ for all model parameters in the warmup stage and then adjusted to $0.01$ and $1e^{-4}$ in the second training stage. 
%
The two training stages both last for $20$k iterations. 
All experiments run on an NVIDIA RTX4090 GPU and last for about $2$ hours.

\myparagraph{Datasets.}
We build the $8$ synthetic dynamic scenes in Table~\ref{tab:nv_ssim} using the Kubric simulator~\citep{Greff2022KubricAS}. Each scene spans $60$ timestamps and contains different numbers of objects in various colors, shapes, and textures. The objects have diverse motion patterns and initial velocities. All images have dimensions of $512 \times 512$ pixels.
We also adopt
real-world scenes from HyperNeRF~\citep{park2021hypernerf} and $\text{D}^2$NeRF~\citep{Wu2022D2NeRFSD}, as shown in Table~\ref{tab:real_world_synthesis} and Table~\ref{tab:real_world_decomposition}.
For the synthetic scenes, we follow D-NeRF~\citep{Pumarola2020DNeRFNR} to employ images collected at viewpoints randomly sampled on the upper hemisphere. In contrast, real-world scenes are captured using a mobile monocular camera.

\begin{itemize}[leftmargin=*]
    \item \textit{3ObjFall.} 
    The scene consists of two cubes and a cylinder. Initially, these objects are positioned randomly within the scene and undergo a free-fall motion along the Z-axis.
    \item
    \textit{3ObjRand.} We use random initial velocities along each object's X and Y axes in \textit{3ObjFall}.
    \item \textit{3ObjMetal.} We change the material of each object in \textit{3ObjFall} from ``Rubber'' to ``Metal''.
    \item \textit{3Fall+3Still.} We add another three static objects with complex geometry to \textit{3ObjFall}.
    \item \textit{6ObjFall} \& \textit{8ObjFall.} We increase the number of objects in \textit{3ObjFall} to $6$ and $8$.
    \item \textit{3ObjRealSimp.} We modify \textit{3ObjFall} with real-world objects with simple textures.
    \item 
    \textit{3ObjRealCmpx.} 
    We modify \textit{3ObjFall} with real-world objects that have complex textures.
    \item \textit{Real-world data.} We adopt the real-world scenes from HyperNeRF~\citep{park2021hypernerf}, $\text{D}^2$NeRF~\citep{Wu2022D2NeRFSD}, and NeRF-DS~\cite{zhiwen2023nerfds}. 
\end{itemize}

\myparagraph{Metrics.}
For novel view synthesis, we report PSNR and SSIM~\citep{Wang2004ImageQA}. 
To compare the scene decomposition results with the 2D methods, we employ the Foreground Adjusted Rand Index (FG-ARI)~\citep{rand1971objective,hubert1985comparing}. 
It measures the similarity of clustering results based on the foreground object masks to the ground truth, ranging from $-1$ to $1$ (higher is better). 
Since no ground truth masks are provided in the real-world datasets, we employ the SAM-Tracker~\cite{cheng2023segment} to annotate one object in each scene. 
Additionally, we include the mean Intersection over Union (mIOU) to provide a comprehensive assessment of the segmentation accuracy and overlap with the ground truth. We utilize the first $5$ frames of the sequence to establish the correspondence between the predicted segmentation and the ground truth. This correspondence is then extrapolated across the entire sequence.

\begin{table*}[t]
\caption{
Scene decomposition results on synthetic datasets. 
To compare with SAVi, which takes as inputs video frames with a fixed viewpoint, we generate synthetic data with a stationary camera and evaluate \newModel{} (\textit{FixCam}) with fixed viewpoints. 
%
}
\label{tab:FG-ARI}
\vspace{-10pt}
\begin{center}
\begin{small}   
\setlength{\tabcolsep}{6pt}{}
\begin{tabular}{lccccccccc}
\toprule
 \multirow{2}{*}{Method}
 &\multirow{2}{*}{3D Decomp.}
 &\multicolumn{2}{c}{3ObjFall} &\multicolumn{2}{c}{3ObjRand} &\multicolumn{2}{c}{3ObjMetal} &\multicolumn{2}{c}{3Fall+3Still} \\
 &&FG-ARI$\uparrow$&mIOU$\uparrow$ &FG-ARI$\uparrow$&mIOU$\uparrow$ &FG-ARI$\uparrow$&mIOU $\uparrow$&FG-ARI$\uparrow$&mIOU$\uparrow$ \\
\cmidrule(lr){1-1}  \cmidrule(lr){2-2} \cmidrule(lr){3-4}  \cmidrule(lr){5-6}  \cmidrule(lr){7-8}  \cmidrule(lr){9-10}
SAVi (\textit{FixCam})  & \XSolidBrush   &3.74 &10.96 &4.38  &11.97  &3.38 & 10.99&6.12  &7.21\\
DynaVol-S (\textit{FixCam}) & \Checkmark
&\textbf{96.02}   &\textbf{94.59}  &\textbf{95.03}   &\textbf{92.66}
 &\textbf{96.61}  &\textbf{95.14}  &\textbf{94.70}    &\textbf{92.98}\\
\cmidrule(lr){1-1}  \cmidrule(lr){2-2} \cmidrule(lr){3-4}  \cmidrule(lr){5-6}  \cmidrule(lr){7-8}  \cmidrule(lr){9-10}
uORF  & \Checkmark          &28.77 &3.67 &38.65 &2.68    &22.58 &3.54 &36.70 &3.15  \\
SAM & \XSolidBrush &70.77  &78.87  &55.52   &65.05  &46.80   &38.59  &47.36  &33.53  \\
MovingParts & \Checkmark &92.44  &21.45  &\underline{97.05}  &55.98  &\underline{95.72} &50.56  &8.37  &11.36 \\
DynaVol & \Checkmark &\underline{96.89} &\underline{90.32}  &96.11  &\underline{66.85}  &85.78  & \underline{59.97} &\underline{92.76}  &\underline{88.20} \\
DynaVol-S & \Checkmark   &\textbf{97.24} &\textbf{91.53} &\textbf{97.23} &\textbf{90.60}  &\textbf{97.25} &\textbf{94.28} &\textbf{95.26}   &\textbf{91.19}   \\

\midrule
\multirow{2}{*}{Method}
&\multirow{2}{*}{3D Decomp.}
&\multicolumn{2}{c}{6ObjFall} &\multicolumn{2}{c}{8ObjFall} &\multicolumn{2}{c}{3ObjRealSimp} &\multicolumn{2}{c}{3ObjRealCmpx} \\
 &&FG-ARI$\uparrow$&mIOU$\uparrow$ &FG-ARI$\uparrow$&mIOU$\uparrow$ &FG-ARI$\uparrow$&mIOU $\uparrow$&FG-ARI$\uparrow$&mIOU$\uparrow$ \\
\cmidrule(lr){1-1}  \cmidrule(lr){2-2} \cmidrule(lr){3-4}  \cmidrule(lr){5-6}  \cmidrule(lr){7-8}  \cmidrule(lr){9-10}
SAVi (\textit{FixCam})  & \XSolidBrush   &6.85   &16.65  &7.87  & 19.44 &3.10  & 14.10 &4.82  &9.98\\
DynaVol-S (\textit{FixCam}) & \Checkmark
&\textbf{94.40} &\textbf{93.04}  &\textbf{94.72}  & \textbf{86.23}  &\textbf{95.54} &\textbf{87.75}   &\textbf{94.34}    & \textbf{92.88}    \\
\cmidrule(lr){1-1}  \cmidrule(lr){2-2} \cmidrule(lr){3-4}  \cmidrule(lr){5-6}  \cmidrule(lr){7-8}  \cmidrule(lr){9-10}
uORF & \Checkmark            &29.23 &3.87  &31.93  & 2.53    &38.26 &2.85 &33.76  &3.39\\
SAM & \XSolidBrush &62.66  &41.35 
&71.68  &28.52  &56.65 &57.46   &51.91  &\underline{75.20} \\
MovingParts & \Checkmark &78.06  &11.79   &89.83  &22.25  &92.34  &36.77  &87.07 &19.91\\
DynaVol & \Checkmark &\underline{95.61} &\underline{89.59} &\underline{93.38} &\textbf{87.16} &\underline{94.28}&\underline{64.75} &\underline{95.02} &63.56\\
DynaVol-S   & \Checkmark       &\textbf{96.08} & \textbf{89.80}&\textbf{95.60} &  \underline{85.41}      &\textbf{95.37} &\textbf{84.55} &\textbf{96.55} &\textbf{92.61} \\
\bottomrule
\end{tabular}
\end{small}
\end{center}
\vspace{-5pt}
\end{table*}


\begin{table*}[t]
\caption{
\revise{3D scene decomposition results on the synthetic datasets in terms of 3D-ARI.}
}
\label{tab:supple_decomp}
\vspace{-10pt}
\begin{center}
\begin{small}   
\setlength{\tabcolsep}{7pt}
\begin{tabular}{lcccccccc}
\toprule
Method &3ObjFall & 3ObjRand & 3ObjMetal & 3Fall+3Still & 6ObjFall &8ObjFall &3ObjRealSimp & 3ObjRealCmpx \\
\midrule
MovingParts &27.02  &37.94  &48.24  &25.52 &27.89 &42.27 &46.12&42.82\\
DynaVol &87.46 &42.10 &66.51 &77.55 &\textbf{88.21} &80.24  &62.25  &63.00\\
\newModel{} &\textbf{90.02}  & \textbf{94.37} &\textbf{87.44}   & \textbf{84.29}  &88.20  & \textbf{81.37} &\textbf{91.00} &\textbf{92.88}\\
\bottomrule
\end{tabular}
\end{small}
\end{center}
\vspace{-5pt}
\end{table*}

\myparagraph{Compared methods.} 
In the novel view synthesis task for synthetic scenes, we evaluate \newModel{} against D-NeRF~\citep{Pumarola2020DNeRFNR},  DeVRF~\citep{Liu2022DeVRFFD}, and TineuVox~\cite{fang2022fast}.
For real-world scenes, we compare it with NeuralDiff~\citep{tschernezki2021neuraldiff}, HyperNeRF~\citep{park2021hypernerf}, TineuVox, and $\text{D}^2$NeRF~\citep{Wu2022D2NeRFSD}. 
Furthermore, we specifically use TineuVox-Small for synthetic scenes and TineuVox-Base for real-world scenes to ensure alignment with our hyperparameter settings.
In the scene decomposition task, we use established 2D/3D object-centric representation learning methods, SAVi~\citep{kipf2021conditional}, uORF~\citep{yu2021unsupervised}, and MovingParts~\cite{yang2024movingparts} as the baselines, where SAVi and uORF are pre-trained on the MOVi-A~\citep{Greff2022KubricAS} and CLEVR-567~\citep{yu2021unsupervised} datasets respectively, which are similar to our synthetic scenes.
Besides, we also include a pretrained and publicly available Segment Anything Model (SAM)~\citep{kirillov2023segany} as a competitive segmentation baseline.
For real-world scenes, we compare \newModel{} with OCLR~\cite{xie2022segmenting} and ClipSeg~\cite{lueddecke22_cvpr}.
%

\begin{table*}[b]
\caption{Scene decomposition results on real-world datasets.
}
\label{tab:real_world_decomposition}
\vspace{-10pt}
\begin{center}
\begin{small}   
\setlength{\tabcolsep}{2pt}{}

\begin{tabular}{lcccccccccccc}
\toprule
& \multicolumn{2}{c}{Chicken} &\multicolumn{2}{c}{Torchocolate} &\multicolumn{2}{c}{Keyboard} &\multicolumn{2}{c}{Split-Cookie} 
&\multicolumn{2}{c}{Americano}
&\multicolumn{2}{c}{Avg.}\\
Method
& FG-ARI$\uparrow$   & mIOU$\uparrow$
& FG-ARI$\uparrow$   & mIOU$\uparrow$
& FG-ARI$\uparrow$   & mIOU$\uparrow$
& FG-ARI$\uparrow$   & mIOU$\uparrow$
& FG-ARI$\uparrow$   & mIOU$\uparrow$
& FG-ARI$\uparrow$   & mIOU$\uparrow$\\
\cmidrule(lr){1-1}  \cmidrule(lr){2-3}  \cmidrule(lr){4-5}  \cmidrule(lr){6-7} 
\cmidrule(lr){8-9} \cmidrule(lr){10-11} \cmidrule(lr){12-13}
OCLR &14.31&14.80 &22.46&20.21 &6.54&10.30  &34.39&34.41 &46.15&35.32 &24.77 &23.01\\
OCLR (\textit{Test Adapt}) &8.48&6.89 &3.06&3.86 &-5.51&1.63 &17.20&14.02 &5.78&3.59  &5.80&5.99\\
ClipSeg &\underline{78.73}&\underline{73.24} &82.98&72.17 &\underline{78.97}&\underline{72.41} &\underline{80.22}&\underline{69.49} &\underline{83.33}&\underline{73.07} &\underline{80.84}&\underline{72.07}\\
DynaVol  &61.92&55.45 &\underline{87.62}&\underline{79.53} &-5.52&0.42 &-4.26&0.11 &83.03&\underline{73.07} &44.55&41.71    \\
DynaVol-S &\textbf{92.91}&\textbf{90.70} &\textbf{91.70}&\textbf{85.56} &\textbf{89.88}&\textbf{85.73} &\textbf{94.49}&\textbf{90.66} &\textbf{95.36}&\textbf{92.11} &\textbf{92.86}&\textbf{88.95}\\
\bottomrule
\end{tabular}
\end{small}
\end{center}
\vspace{-10pt}
\end{table*}


\myparagraph{Scene decomposition.} 
To get the 2D segmentation results, we assign the rays to different slots according to the contribution of each slot to the final color of the ray. 
Specifically, suppose $p_{in}$ is the occupancy probability of latent code $n$ at point $i$, we have
\begin{equation}
    \widehat{P}_{n}(\mathbf{r}) = \sum_{i=1}^P T_i(1-\exp(-\sigma_i \delta_i))p_{i,n},
\end{equation}
where $p_{in}$ is the corresponding occupancy probability and $\widehat{P}_{n}(\mathbf{r})$ is the color contribution to the final color of latent code $n$. We can then predict the label of 2D segmentation by $\hat{y}(\mathbf{r}) = \mathrm{argmax}_n(\widehat{P}_{n}(\mathbf{r}))$.

\myparagraph{Dynamic scene editing.} The object-centric representations acquired through \newModel{} demonstrate the capability for seamless integration into scene editing workflows, eliminating the need for additional training. 
By manipulating the 4D voxel grid $\major{\mathcal{V}^\text{Opac}}
$ learned in \newModel{}, objects can be replaced, removed, duplicated, or added according to specific requirements. 
For example, we can swap the color between objects by swapping the corresponding latent codes assigned to each object.
Moreover, the deformation field of individual objects can be replaced with user-defined trajectories (\textit{e.g.}, rotations and translations), enabling precise animation of the objects in the scene.

\subsection{Novel View Synthesis}

We compare the performance of \newModel{} for novel view synthesis with D-NeRF, DeVRF, and TineuVox.
%
%
As DeVRF is trained with $V=60$ views at the initial timestamp and $4$ views for subsequent timestamps, we also train a DeVRF model under a data configuration similar to ours. Specifically, we only have access to one image per timestamp ($t \ge 1$) in a dynamically sampled view from the upper hemisphere. This model is termed as ``\textit{DeVRF-Dyn}''. 
As shown in Table~\ref{tab:nv_ssim}, \textit{DeVRF-Dyn} yields a significant decline in performance compared to its standard baseline, primarily due to its heavy reliance on accurate initial scene understanding. In contrast, \newModel{} achieves the best results across most scenes among all the compared models.

We also evaluate the performance of \newModel{} in real-world scenes with cutting-edge NeRF-based methods, including NeurDiff, HyperNeRF, TineuVox, and $\text{D}^2$NeRF. 
As shown in Table~\ref{tab:real_world_synthesis}, \newModel{} outperforms the previous state-of-the-art method, HyperNeRF, by $10.8\%$ in PSNR and $8.3\%$ in SSIM on average. 
Furthermore, we also compare \newModel{} with another neural rendering method that learns disentangled representations. The results indicate that our approach outperforms $\text{D}^2$NeRF by $8.2\%$ in PSNR and $3.5\%$ in SSIM.

\begin{figure*}[t]
    \centering
    \includegraphics[width=0.96\textwidth]{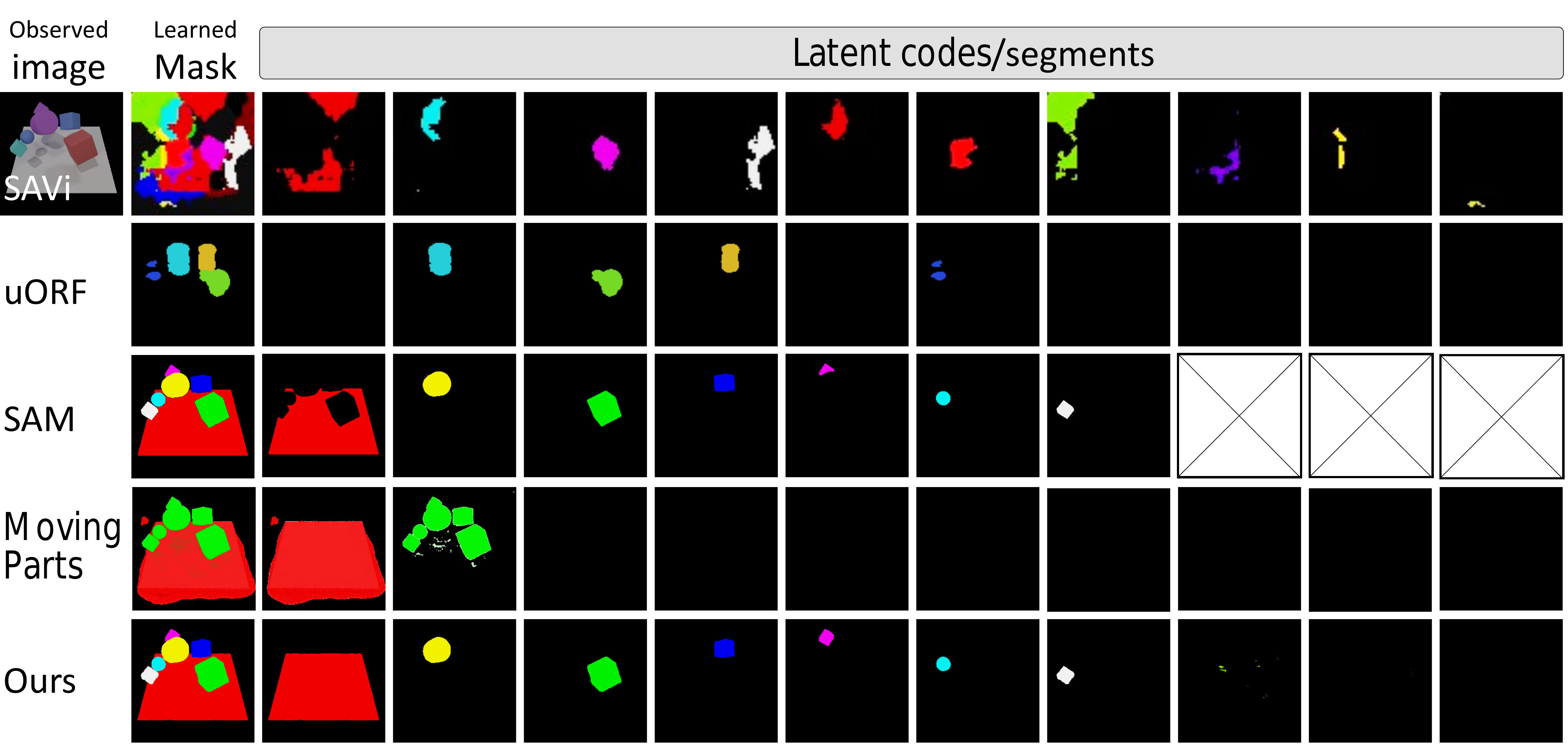}
    \vspace{-5pt}
    \caption{Visualization of scene decomposition results for each object in the synthetic \textit{6ObjFall} scene.}
\label{fig:compare_seg}
\vspace{-5pt}
\end{figure*}

\begin{figure*}[!t]
    \centering
    \includegraphics[width=0.97\textwidth]{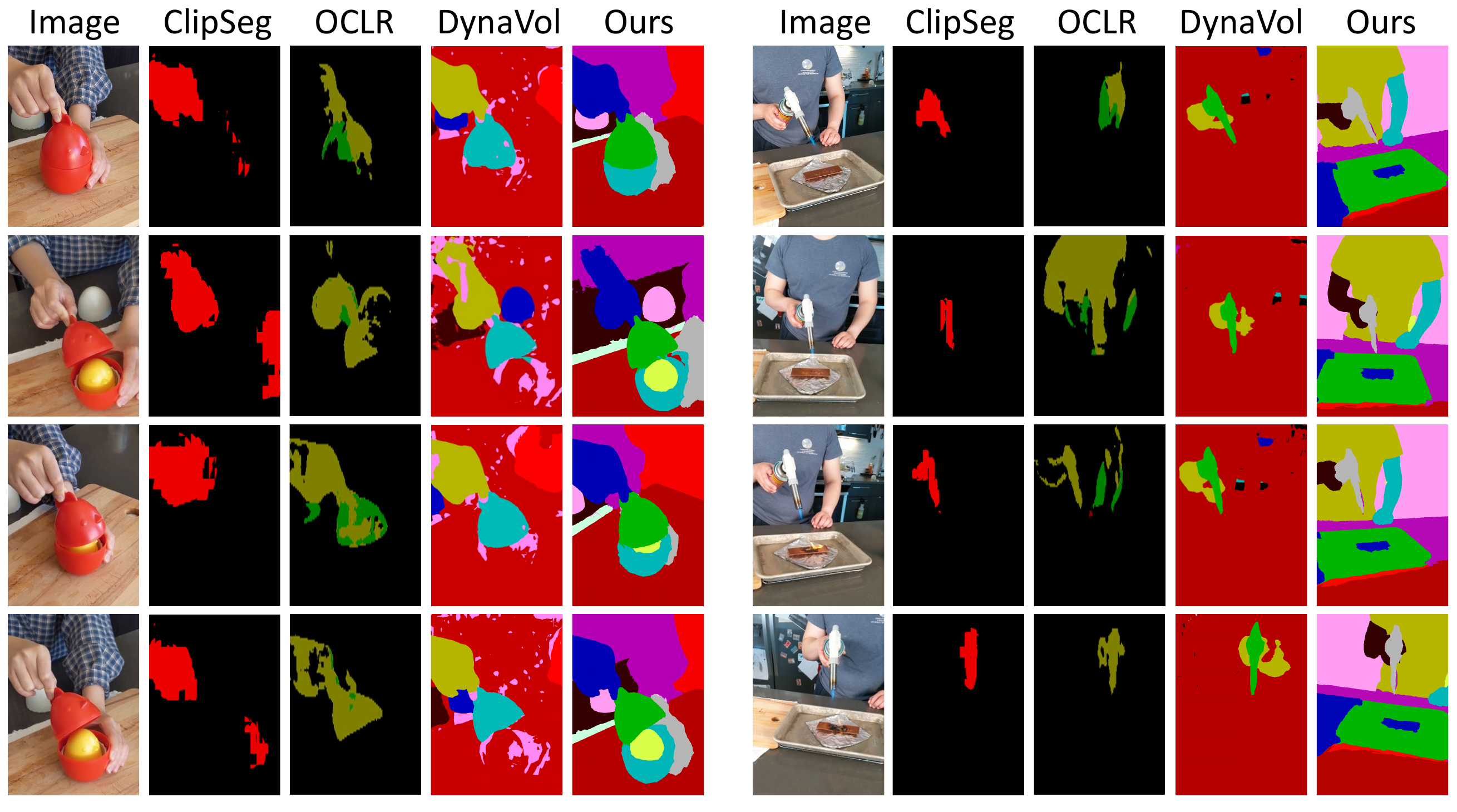}
    \vspace{-5pt}
    \caption{Real-world scene decomposition results for \textit{Chicken} and \textit{Torchocolates}.}
\label{fig:seg_realworld}
\vspace{-10pt}
\end{figure*}

\begin{figure*}[t]
    \centering
    \includegraphics[width=0.99\textwidth]{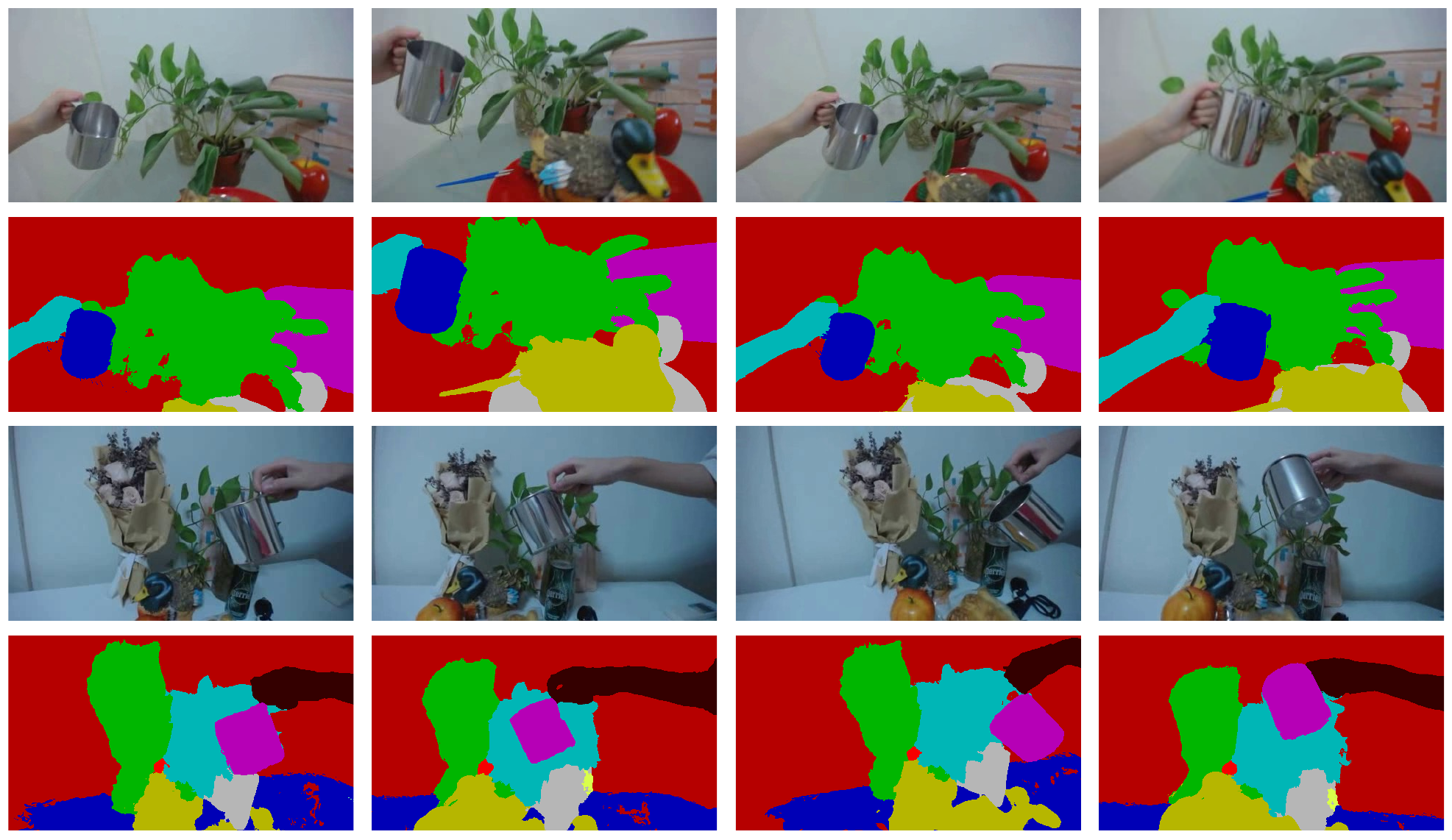}
    \vspace{-5pt}
    \caption{Real-world scene decomposition results for \textit{Cup} and \textit{Sieve}.}
    \label{fig:seg_realworld2}
    \vspace{-5pt}
\end{figure*}

Fig.~\ref{fig:novel_view} showcases the rendered images at an arbitrary timestamp from a novel view.
On the synthetic dataset, it shows that \newModel{} captures 3D geometries and the motion patterns of different objects more accurately than the compared methods.
In contrast, D-NeRF fails to model the complex motion in \textit{3ObjRand} and generates blurry results in \textit{3ObjRealCpmx}.
DeVRF, on the other hand, also fails to accurately model the trajectories of the moving object with complex textures, as highlighted by the red box in \textit{3ObjRealCmpx}.
The real-world examples on the right side in Fig.~\ref{fig:novel_view} illustrate that \newModel{} generates more high-quality novel view images than the prior art, TineuVox.

\subsection{Scene Decomposition}

To obtain quantitative results for 2D segmentation as well as the scene decomposition maps, we assign the casted rays in volume rendering to different latent codes according to each latent code's contribution to the ray's final color. 
Table~\ref{tab:FG-ARI} provides a comparison between \newModel{} with 2D/3D unsupervised object-centric decomposition methods (SAVi, uORF, and MovingParts) and a pretrained SAM.  
For a fair comparison with SAVi, which works on 2D video inputs, we implement a \newModel{} model using consecutive images with a fixed camera view (termed as \textit{FixCam}). 
For uORF and SAM, as they are primarily designed for static scenes, we preprocess the dynamic sequence into $T$ individual static scenes as the inputs of these models and evaluate their average performance on the whole sequence.
For MovingParts, which generates hierarchical clustering results by iteratively merging the slots with the most similar motions, we select the outcome that achieves the highest mIOU as its final result.
The FG-ARI and mIOU results in Table~\ref{tab:FG-ARI} show that \newModel{} outperform all of the compared models by large margins.

\revise{Furthermore, we assess scene decomposition in 3D space. Specifically, we first voxelize the ground truth mesh for each object in the synthetic scenes to obtain their surface points. These surface points are then used to query the corresponding predicted labels in $\mathcal{V}^\text{Occ}$. The predicted labels are compared to the ground truth by calculating the 3D-ARI for each scene, as presented in Table~\ref{tab:supple_decomp}. The results demonstrate that \newModel{} consistently outperforms other baselines across all synthetic datasets.}

In Fig.~\ref{fig:compare_seg}, we randomly select a timestamp on 6ObjFall and present the object-centric decomposition maps of SAVi, uORF, SAM, MovingParts, and \newModel{}. 
We have the following three observations from the visualized examples.
First, compared with the 2D models (SAVi and SAM), \newModel{} can effectively handle severe occlusions between objects in 3D space. It shows the ability to infer the complete shape of the objects, as illustrated in the $1$st and $5$th latent codes.  
Second, compared with the 3D decomposition uORF, \newModel{} can better segment the dynamic scene by leveraging explicitly meaningful spatiotemporal representations, while uORF only learns latent representations for each object. 
Furthermore, \newModel{} is shown to adaptively work with redundant latent codes, in the sense that the pre-defined number of latent codes can be larger than the actual number of objects in the scene. In such cases, the additional latent codes learn to disentangle noise in visual observations or learn to not contribute significantly to image rendering, enhancing the model's flexibility and robustness.
\revise{We further provide mesh visualization in Sec.~\ref{analysis} to demonstrate the ability of our approach for disentanglement representation learning.}

We further evaluate the scene decomposition results of \newModel{} in real-world scenes against the state-of-the-art unsupervised video segmentation method, OCLR~\cite{xie2022segmenting}. 
Notably, OCLR utilizes DINOv1 features~\citep{caron2021emerging} for test-time adaptation, referred to as OCLR (\textit{Test Adapt}), to enhance temporal consistency. 
Additionally, we compare \newModel{} with the open-world object segmentation approach, ClipSeg~\cite{lueddecke22_cvpr}. 
Since ClipSeg relies on text prompts, we manually adjust the prompts to tune the model. 
It is important to highlight that ClipSeg employs the CLIP model~\cite{radford2021learning} and is further trained with true segmentation in a supervised manner, whereas our approach is entirely unsupervised. 
As demonstrated in Table~\ref{tab:real_world_decomposition}, \newModel{} consistently achieves the best decomposition results, significantly outperforming all compared methods.

\begin{figure*}[t]
    \centering
    \includegraphics[width=\textwidth]{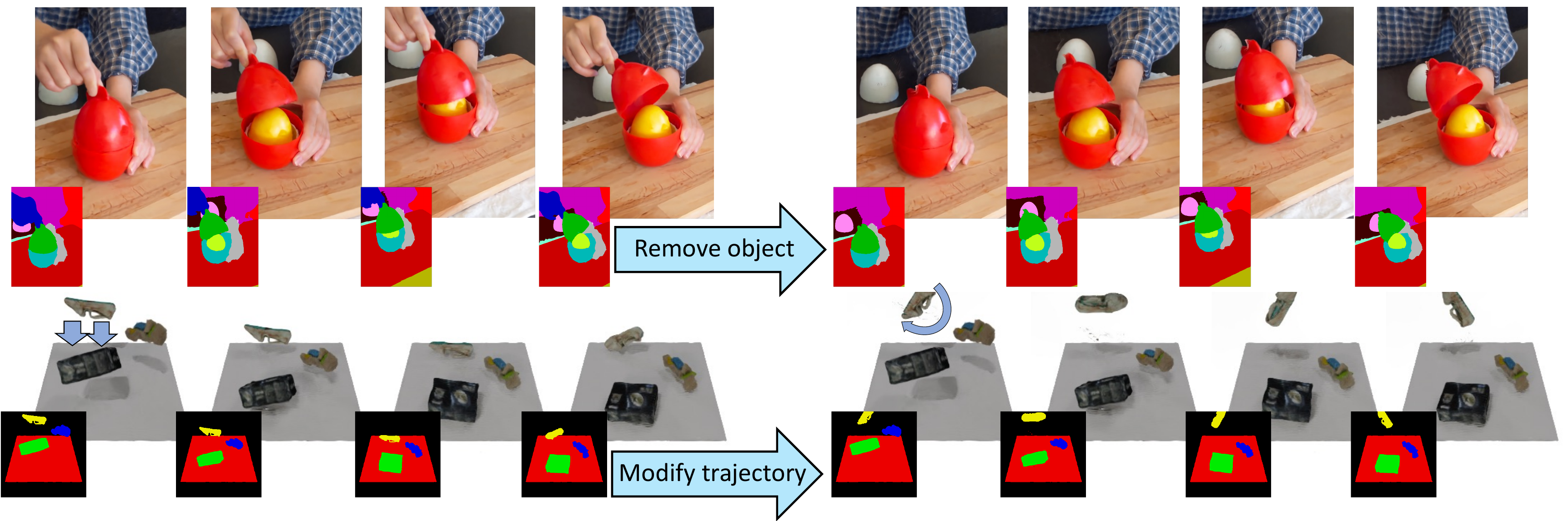}
    \vspace{-15pt}
    \caption{Dynamic scene editing examples for real-world (top) and synthetic scenes (bottom). We optimize \newModel{} on the original scene, manually edit its object-centric voxel grids (\textit{e.g.}, object removal and trajectory modification), and subsequently generate a new image sequence based on the edited voxel grids.
    }
    \label{fig:edit}
\end{figure*}

\begin{table*}[t]
\caption{
Ablation studies of the key components and the training stages in \newModel{}. \revise{Additionally, we showcase the effect of using alternative designs for volume slot attention in Fig.~\ref{fig:volume_sa}.}
}
\vspace{-10pt}
\label{tab:ablation:modules}
\begin{center}
\begin{small}   
\setlength{\tabcolsep}{15pt}{}
\begin{tabular}{lcccccc}
\toprule
\multirow{2}{*}{Method} & \multicolumn{2}{c}{3ObjRand} &\multicolumn{2}{c}{8ObjFall} &\multicolumn{2}{c}{3ObjRealCmpx} \\
& PSNR$\uparrow$   & mIOU$\uparrow$
& PSNR$\uparrow$  & mIOU$\uparrow$
& PSNR$\uparrow$   & mIOU$\uparrow$\\
\cmidrule(lr){1-1}  \cmidrule(lr){2-3}  \cmidrule(lr){4-5}  \cmidrule(lr){6-7}
\major{w/o Object-centric voxelization} &32.15 &N/A &30.76 &N/A &28.79 &N/A\\
w/o Forward deformation $f_{\xi}^{\prime}(\cdot)$      &32.15&90.49  &31.55&75.91&\textbf{29.05}&\textbf{92.78}\\
w/o CRF       &\underline{32.57}&\underline{90.56}&\textbf{31.79}&\underline{85.37}&28.85&92.49\\
w/o Multi-grids joint optim.  &30.48&90.24  &29.03&84.46&26.81&92.37 \\
Multi-grids joint optim. from scratch &28.86&14.73&29.01&5.82&25.64& 11.09\\
\cmidrule(lr){1-1}  \cmidrule(lr){2-3}  \cmidrule(lr){4-5}  \cmidrule(lr){6-7}
Full model      &\textbf{32.58}&\textbf{90.60}    &\underline{31.78}&\textbf{85.41} 
&\underline{28.82}&\underline{92.61}  \\
\bottomrule
\end{tabular}
\end{small}
\end{center}
\vspace{-10pt}
\end{table*}

In Fig.~\ref{fig:seg_realworld}, we showcase the real-world scene decomposition results of \newModel{} 
compared with OCLR and ClipSeg. 
Our findings reveal that \newModel{} produces clearer 3D segmentation for each object, especially at object boundaries, while also maintaining temporal consistency across the video frames. 
In contrast, OCLR struggles to preserve temporal consistency. \newModel{} also outperforms our previous work, \model{}, indicating the importance and effectiveness of incorporating semantic features.

\revise{
Can \newModel{} handle scenarios where objects leave or new objects enter the scene? The answer is yes. 
For example, in the left case in Fig.~\ref{fig:seg_realworld}, the inner part of the \textit{Chicken} toy (the yellow one) repeatedly leaves and reappears throughout the sequence, yet \newModel{} successfully generates consistent labels. 
This robustness is primarily due to \newModel{} performing segmentation in the canonical space, where all objects appearing in the sequence are represented, regardless of whether they are visible in specific frames.}
More real-world examples with severe occlusions can be found in Fig.~\ref{fig:seg_realworld2} and supplementary materials.

\begin{table}[t]
\vspace{4pt}
\caption{
The analysis of the impact of the number of latent codes ($N$) on the performance of our approach.
}
\label{tab:ablation:slots_num}
\vspace{-10pt}
\begin{center}
\begin{small}   
\setlength{\tabcolsep}{5pt}{}
\begin{tabular}{lcccccc}
\toprule
\multirow{2}{*}{$N$} & \multicolumn{2}{c}{3ObjRand} &\multicolumn{2}{c}{8ObjFall} &\multicolumn{2}{c}{3ObjRealCmpx} \\

 & PSNR   & mIOU
& PSNR  & mIOU
& PSNR   & mIOU
\\
\cmidrule(lr){1-1}  \cmidrule(lr){2-3}  \cmidrule(lr){4-5}  \cmidrule(lr){6-7} 
5     &32.53&\textbf{90.69}    &31.73&42.80&\underline{28.82}&\underline{92.60}  \\
10 (Final)  &\underline{32.58}&90.60    &\underline{31.78}&\underline{85.41} 
&\underline{28.82}&\textbf{92.61}  \\
15 &\textbf{32.59}&\underline{90.67}&\textbf{31.79}&\textbf{85.46}&\textbf{28.86}&92.59 \\
\bottomrule
\end{tabular}
\end{small}
\end{center}
\vspace{-5pt}
\end{table}

\subsection{Dynamic Scene Editing}

After the training period, the object-centric voxel representations learned by \newModel{} can be readily used in downstream tasks such as scene editing without the need for additional model tuning. \newModel{} allows for easy manipulation of the observed scene by directly modifying the object occupancy values within the voxel grids or switching the learned deformation function to a pre-defined one. This flexibility empowers users to make various scene edits and modifications. For instance, in the first example in Fig.~\ref{fig:edit}, we remove the hand that is pinching the toys. In the second example, we modify the dynamics of the shoe from falling to rotating. More showcases are included in the supplementary materials.


\subsection{Model Analyses}
\label{analysis}
\myparagraph{Ablation studies.}
Table~\ref{tab:ablation:modules} presents the ablation study results, highlighting the effectiveness of the key component of \newModel{}.
\revise{First, we observe that object-centric voxelization enables effective scene decomposition and improves the rendering results.}
Subsequently, we can find that the forward deformation network $f_{\xi}^{\prime}(\cdot)$ is crucial to the final rendering and decomposition results. 
%
%
Furthermore, in the absence of the multi-grids joint optimization stage, the performance of \newModel{} significantly degrades, highlighting the importance of refining the object-centric voxel representation with a latent-code-based renderer.
Additionally, we conduct an ablation study that performs multi-grids joint optimization from scratch with randomly initialized $\major{\mathcal{V}^\text{Occ}}$ and $f_\phi(\cdot)$. The results clearly show that excluding the warmup stage substantially impacts the final performance, especially for the scene decomposition results.
\myparagraph{The number of object-centric latent codes.}
First, we examine the impact of the number of latent codes ($N$), which indicates the maximum number of objects in the scene. 
According to Table~\ref{tab:ablation:slots_num}, the model achieves comparable results as long as $N$ exceeds the number of objects in the scene. The presence of redundant latent codes has only a minor influence on both rendering and decomposition results.
Additionally, we showcase the per-object rendering results in supplementary materials, demonstrating \newModel{}'s robustness to variations in the number of latent codes. Our findings reveal that redundant latent codes typically remain unassigned, effectively addressing the potential issue of over-segmentation.

\begin{table}[t]
\vspace{4pt}
\caption{
Hyperparameter analyses of (1) the threshold value for splitting the foreground and background in the object-centric voxel grids initialization stage, and (2) the weight of the per-point RGB loss ($\alpha_p$). We use $\delta_\text{den}=1, \alpha_p=0.01$ for the final model.
}
\label{tab:ablation:thresh}
\vspace{-10pt}
\begin{center}
\begin{small}   
\setlength{\tabcolsep}{3pt}{}
\begin{tabular}{lcccccc}
\toprule
\multirow{2}{*}{Hyperparameter}& \multicolumn{2}{c}{3ObjRand} &\multicolumn{2}{c}{8ObjFall} &\multicolumn{2}{c}{3ObjRealCmpx} \\

& PSNR   & mIOU
& PSNR  & mIOU
& PSNR  & mIOU
\\
\cmidrule(lr){1-1}  \cmidrule(lr){2-3}  \cmidrule(lr){4-5}  \cmidrule(lr){6-7} 

$\delta_\text{den}=0.5$  &32.56 &\textbf{90.71} &\underline{31.77}&85.20 &28.82&\textbf{92.61}     \\
$\delta_\text{den}=2$ &\underline{32.57}&\underline{90.65} &31.74&\textbf{88.41} &\underline{28.83}&\textbf{92.61}      \\
\cmidrule(lr){1-1}  \cmidrule(lr){2-3}  \cmidrule(lr){4-5}  \cmidrule(lr){6-7} 
$\alpha_p=0$ &30.00&88.66 &30.64&68.31 &28.16&\underline{92.45}\\
$\alpha_p = 0.1$  &30.95&31.61  &31.15&23.89 &\textbf{28.89}&84.25 \\
\cmidrule(lr){1-1}  \cmidrule(lr){2-3}  \cmidrule(lr){4-5}  \cmidrule(lr){6-7} 
 Final model   &\textbf{32.58}&90.60    &\textbf{31.78}&\underline{85.41} 
&28.82&\textbf{92.61}  \\
\bottomrule
\end{tabular}
\end{small}
\end{center}
\vspace{-15pt}
\end{table}

%
%


\begin{figure*}[t]
    \centering
    \includegraphics[width=0.95\textwidth]{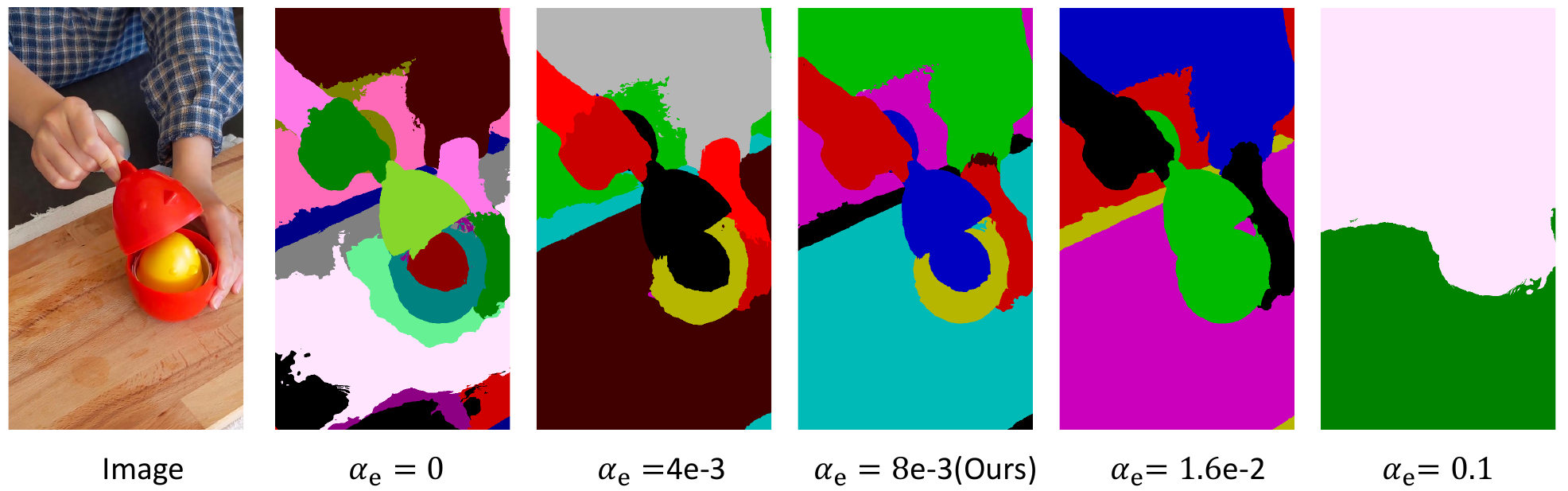}
    \vspace{-5pt}
    \caption{Visualization of semantic probabilities learned with different weights of the entropy loss ($\alpha_e$).}
\label{fig:entropy_loss}
\vspace{-5pt}
\end{figure*}

\begin{figure*}[t]
    \centering
    \includegraphics[width=0.95\textwidth]{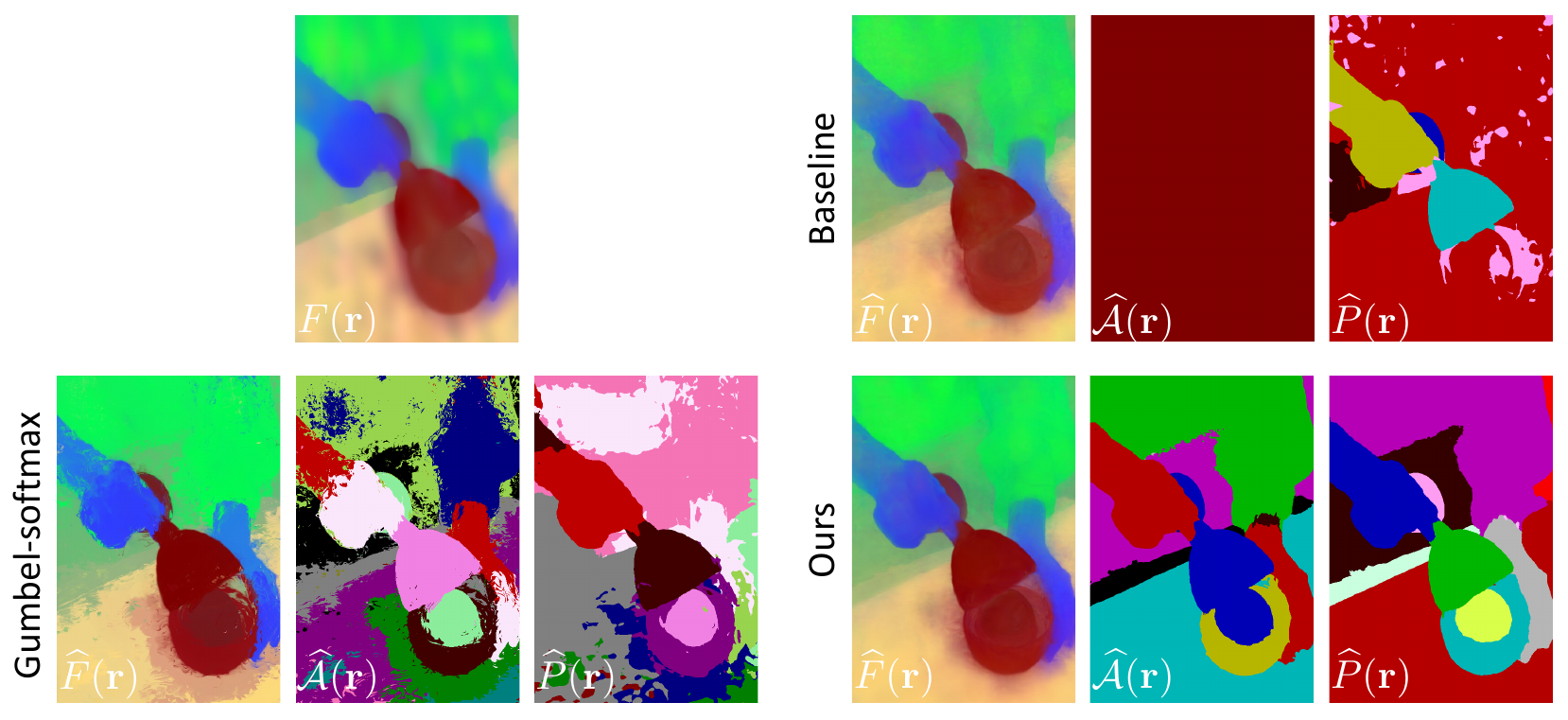}
    \vspace{-5pt}
    \caption{
    Qualitative results of alternative designs for the slot attention module.
    From left to right: Visualization of reconstructed semantic features $\widehat{F}(\mathbf{r})$, corresponding semantic probabilities $\widehat{\mathcal{A}}(\mathbf{r})$, and scene decomposition results $\widehat{P}(\mathbf{r})$ for various design choices of the volume slot attention module. $F(\mathbf{r})$ denotes the 2D semantic features extracted by DINOv2. 
    }
\label{fig:volume_sa}
\vspace{-5pt}
\end{figure*}

\myparagraph{Hyperparameters for the loss functions.} 
The hyperparameters in the loss functions are set to $\alpha_p=0.01$, $\alpha_b=0.01$, $\alpha_e=0.008$, and $\alpha_c=1.0$.
We study the hyperparameter choice of the threshold value for splitting the foreground and background in the warmup stage, and the weight ($\alpha_p$) of per-point RGB loss in Table~\ref{tab:ablation:thresh}.
We find that the performance of \newModel{} remains robust to different threshold values. 
Furthermore, $\alpha_p=0.01$ performs better than $\alpha_p=0.0$, indicating that a conservative use of $\mathcal{L}_\text{point}$ can improve the performance. 
A possible reason is that it eases the training process by moderately penalizing the discrepancy of nearby sampling points on the same ray.
When the weight of $\mathcal{L}_\text{point}$ increases to $\alpha_p=0.1$, the method's performance decreases significantly. 
It is because too much emphasis on $\mathcal{L}_\text{point}$ is intuitively unreasonable and can potentially introduce bias to the neural rendering process.
The entropy loss in Eq.~\eqref{eq:semantic_loss} encourages the semantic probabilities $\widehat{\mathcal{A}}$ learned in the warmup phase to approach a one-hot distribution. 
In Fig.~\ref{fig:entropy_loss}, we visualize $\widehat{\mathcal{A}}$ learned with different values of $\alpha_e$. 
By increasing $\alpha_e$, we prevent the model from over-segmenting the 3D scene with a redundant number of slot features. In other words, tuning $\alpha_e$ allows us to use a larger default number of $N$ while balancing between over-segmentation and under-segmentation. 

\begin{figure*}[t]
    \centering
    \includegraphics[width=0.9\textwidth]{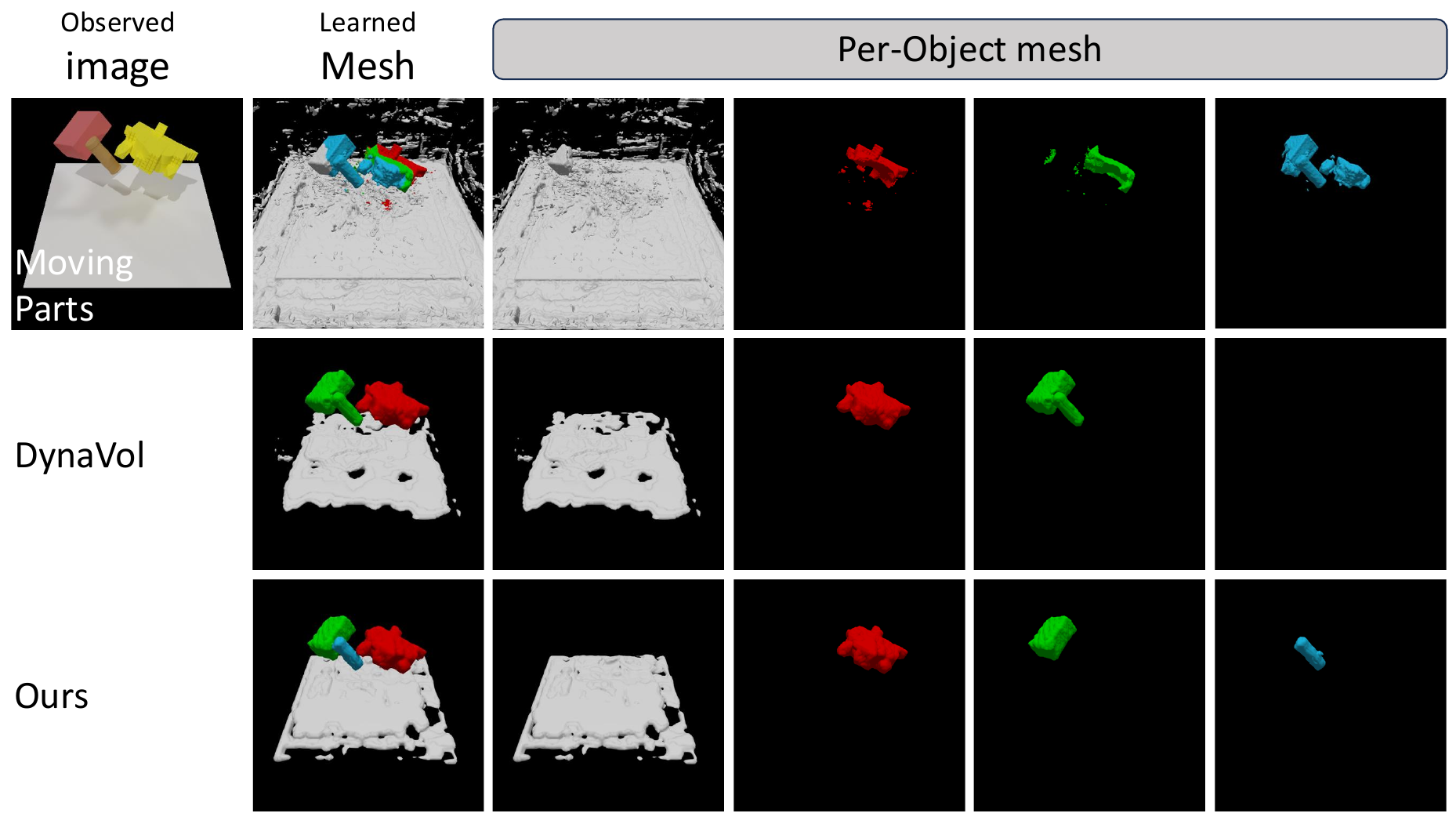}
    \vspace{-5pt}
    \caption{\revise{Visualization of per-object mesh reconstruction results on \textit{3ObjRealSimp}, with each object denoted by a different color.}}
\label{fig:mesh}
\end{figure*}

\begin{table*}[t]
\caption{
The analysis of using different semantic feature extractors.
}
\label{tab:ablation:feature_extractor}
\vspace{-10pt}
\begin{center}
\begin{small}   
\setlength{\tabcolsep}{9pt}{}
\begin{tabular}{lcccccccccc}
\toprule
\multirow{2}{*}{Semantic Feature Extractor}& \multicolumn{2}{c}{Chicken} &\multicolumn{2}{c}{Torchocolate} &\multicolumn{2}{c}{Keyboard} &\multicolumn{2}{c}{Avg.}  \\

 & FG-ARI$\uparrow$   & mIOU$\uparrow$
& FG-ARI$\uparrow$  & mIOU$\uparrow$
& FG-ARI$\uparrow$   & mIOU$\uparrow$
& FG-ARI$\uparrow$   & mIOU$\uparrow$
\\
\cmidrule(lr){1-1}  \cmidrule(lr){2-3}  \cmidrule(lr){4-5}  \cmidrule(lr){6-7}  \cmidrule(lr){8-9} 
None (\model{}) &61.92&55.45 &\underline{87.62}&\underline{79.53} &-5.52&0.42 &48.01 &45.13\\
DINOv1 &91.33&88.76 &6.40&7.85 &\underline{87.83}&\underline{83.18}  &\underline{61.85}&59.93\\
MaskClip &24.94&36.79 &-1.64&0.28 &-8.90&14.00 &4.80&17.02\\
ViT &\textbf{95.84}&\textbf{94.59} &3.37&5.91 &86.52&81.70 &61.01&\underline{60.73}\\
DINOv2 (in \newModel{})  &\underline{92.91}&\underline{90.70} &\textbf{91.70}&\textbf{85.56} &\textbf{89.88}&\textbf{85.73} &\textbf{91.49}&\textbf{87.33} \\
\bottomrule
\end{tabular}
\end{small}
\end{center}
\vspace{-10pt}
\end{table*}

\begin{table}[t]
\vspace{4pt}
\caption{
\revise{The analysis of model efficiency on real-world datasets and resource requirements of each training stage of our model.}
}
\label{tab:ablation:efficiency}
\vspace{-10pt}
\begin{center}
\begin{small}   
\setlength{\tabcolsep}{7pt}{}
\begin{tabular}{lcccccccccc}
\toprule
Method &  Train (hour) & Memory (GB) & Test (FPS) \\
\midrule
HyperNeRF &32  & 128 &0.04\\
$\text{D}^2$NeRF  &8 &320 & 0.04\\
TineuVox &0.5 &22 & 0.22 \\
DynaVol & 3 &16 & 0.25 \\
DynaVol-S &1.8 &21  &0.28\\
\midrule
Stage 1 & 1&21 & N/A \\
Stage 2 &0.3 &N/A & N/A \\
Stage 3 &0.5 &14 & N/A \\
\bottomrule
\end{tabular}
\end{small}
\end{center}
\vspace{-15pt}
\end{table}

\begin{table*}[b]
\vspace{-5pt}
\caption{
\revise{Analyses of different training strategies for the bi-directional deformation networks.}
}
\label{tab:supple_dynamics}
\vspace{-10pt}
\begin{center}
\begin{small}   
\setlength{\tabcolsep}{7pt}
\begin{tabular}{lcccccccccccc}
\toprule
\multirow{2}{*}{Method} & \multicolumn{2}{c}{3ObjRand} & \multicolumn{2}{c}{8ObjFall} & \multicolumn{2}{c}{3ObjRealCmpx}& \multicolumn{2}{c}{Chicken} & \multicolumn{2}{c}{Average} \\

 & PSNR$\uparrow$ & mIOU$\uparrow$ & PSNR$\uparrow$ & mIOU$\uparrow$ & PSNR$\uparrow$ & mIOU$\uparrow$ & PSNR$\uparrow$ & mIOU$\uparrow$ &PSNR$\uparrow$ & mIOU$\uparrow$ \\
\cmidrule(lr){1-1} \cmidrule(lr){2-3} \cmidrule(lr){4-5} \cmidrule(lr){6-7} \cmidrule(lr){8-9}  \cmidrule(lr){10-11}
Separate Training &32.45 &90.60 &31.92 &86.43 &28.86 &92.61 &28.44 &90.55 &30.41&90.05 \\
Co-Training &32.58&90.60    &31.78&85.41 
&28.82&92.61  &28.43&90.70 &30.40&89.83\\
Stop-gradient &32.23 &80.34  &30.32 &69.63  &28.77&92.60  &28.44 &90.74   &29.94 &83.33  \\
\bottomrule
\end{tabular}
\end{small}
\end{center}
\end{table*}

\myparagraph{Alternative designs for volume slot attention.}
In considering the design of the volume slot attention module, we perform the experiments shown in Fig.~\ref{fig:volume_sa}.
A straightforward approach, referred to as ``Baseline'', rewrites Eq.~\eqref{eq:semantic_attn} by removing the slot attention operator, such that: $\mathbf{f}_i^s = \text{Interp}\left(\mathbf{x}_i+f_\psi(\mathbf{x}_i, t),\major{\mathcal{V}^\text{Sem}}\right)$ and $\widehat{F}(\mathbf{r}) = \sum_{i=1}^P T_i\left(1-\exp(-\sigma_i\delta_i)\right)\mathbf{f}_{i}^s$.
%
In the Baseline model, we otherwise obtain $\mathcal{A}_i$ by applying the connected components algorithm to the cosine similarities greater than $0.9$ between adjacent voxel features.
We can find that the Baseline model tends to merge features into a single cluster in $\widehat{\mathcal{A}}$.
As shown, the ambiguous estimation of $\widehat{\mathcal{A}}$ results in a noisy outcome for the learned semantic features, thereby undermining the final scene decomposition results presented by \major{$\mathcal{V}^\text{Occ}$}.
%
Another alternative approach uses the Gumbel-softmax to replace the original softmax operation in Eq.~\eqref{eq:semantic_attn}, which ensures that each voxel is exclusively associated with a single slot, with $\mathcal{A}_i$ representing a one-hot vector.
In Fig.~\ref{fig:volume_sa}, we visualize the generated semantic features $\widehat{F}(\mathbf{r})$, the corresponding semantic probabilities $\widehat{\mathcal{A}}(\mathbf{r})$ obtained from Eq.~\eqref{eq:sem_feature}, and the scene decomposition results $\widehat{P}(\mathbf{r})$. 
As we can see, the Gumbel-softmax method leads to noisy decomposition results due to its hard assignment strategy.
Our method, in contrast, effectively mitigates the overfitting of 2D semantic feature noise, resulting in clearer boundaries and more accurate decomposition outcomes.
It is important to note that $\mathcal{A}_i$ plays a crucial role in initializing 4D voxel grids, $\major{\mathcal{V}^\text{Occ}}$, which in turn affects the final decomposition results.
Specifically, if $\mathcal{A}_i$ merges features excessively, \newModel{} will lose the insights provided by the semantic features, potentially degrading to \model{}.

\myparagraph{Alternative semantic feature extractors.} 
In addition to DINOv2, our approach can be seamlessly integrated with other semantic feature extractors, such as MaskClip~\cite{dong2023maskclip}, ViT~\cite{dosovitskiy2020image}, and DINOv1~\cite{caron2021emerging}. 
In Table~\ref{tab:ablation:feature_extractor}, we explore the impact of these models. 
The results show that DINOv2 outperforms the other models. 
Aside from feature quality, these discrepancies may also stem from DINOv2's strong capabilities in maintaining multi-view consistency~\cite{el2024probing}.
\revise{In the supplementary materials, we further explore the impact of using different variants of the DINOv2 model on the final performance of 3D decomposition, particularly for objects with intricate shapes.}

\myparagraph{Visualized per-object mesh reconstruction.}
\revise{To provide an intuitive understanding of its decomposition performance, we use the learned occupancy field to reconstruct the mesh for each object. As shown in Fig.~\ref{fig:mesh}, \newModel{} demonstrates significant improvements over the baselines, including MovingParts, in both the accuracy of reconstructed geometries and decomposition results.}

\myparagraph{Model efficiency.}
\revise{In Table~\ref{tab:ablation:efficiency}, we compare \newModel{} with existing approaches in terms of training time, memory consumption, and rendering FPS on real-world data. Additionally, we provide a breakdown of the resource requirements for each training stage of our model.
The results demonstrate that \newModel{} achieves comparable training efficiency to the baseline models.
Furthermore, due to our explicit opacity representation, \newModel{} significantly reduces computational costs by filtering out low-opacity points, leading to the highest rendering FPS and reduced memory usage across all evaluated methods.
Although \newModel{} requires a longer training time than TineuVox, the primary overhead arises from training the occupancy and semantic voxel grids, $\mathcal{V}^\text{Occ}$ and $\mathcal{V}^\text{Sem}$, which facilitate scene decomposition—a capability absent in other methods.
A potential direction for future research is to explore the use of K-Planes~\cite{fridovich2023k} to further minimize training time and memory consumption.
}

\myparagraph{Analyses of bi-directional deformation networks.}
\revise{As discussed in Sec.~\ref{sec:warm}, training the forward and the backward deformation networks simultaneously may raise concerns about whether both networks can be properly trained.
To address this issue, we implement two alternative training strategies: ``\textit{Separate}'' and ``\textit{Stop-gradient}''. In the \textit{separate} strategy, we first use the image reconstruction loss to train the backward deformation network $f_\psi(\cdot)$, then use the fixed $f_\psi(\cdot)$ to supervise the forward modeling network $f^\prime_\xi(\cdot)$, ensuring both networks are well-initialized before joint training. The ``\textit{Stop-gradient}'' strategy involves applying stop gradients to either side of Eq.~\eqref{eq:cyc-loss}, such that:
\begin{equation}
\begin{split}
    \mathcal{L}_\text{Cyc} = \frac{1}{|\mathcal{R}|}\sum_{{r}\in\mathcal{R}}\frac{1}{P}\sum_{i=1}^{P} \Big(  & \left\| \mathrm{sg}(f_{\psi} (x_i,t)) + f^\prime_{\xi} (x^\prime_i,t)\right\|_2^2 \\
    + & \left\| f_{\psi} (x_i,t) +  \mathrm{sg}(f^\prime_{\xi} (x^\prime_i,t))\right\|_2^2\Big).
\end{split}
\end{equation}
As shown in Table~\ref{tab:supple_dynamics}, the \textit{separate training} method yields comparable performance to the proposed co-training method, while the \textit{stop-gradient} method degrades model performance, particularly in synthetic scenes. We adopt the co-training strategy in DynaVol-S for its balance of performance and simplicity.}

%% file: text/related.tex
\section{Related Work}

\myparagraph{Unsupervised 2D scene decomposition.}
Most existing methods in this area~\citep{greff2016tagger,greff2019multi,burgess2019monet,engelcke2019genesis} use latent features to represent objects in 2D scenes.
The slot attention method~\citep{locatello2020object} extracts object-centric latents with an attention block and repeatedly refines them using GRUs~\citep{Cho2014LearningPR}.
SAVi~\citep{kipf2021conditional} extends slot attention to dynamic scenes by updating slots at each frame and using optical flow as the training target. 
STEVE~\citep{singh2022simple} improves SAVi by replacing its spatial broadcast decoder with an autoregressive Transformer.
SAVi++~\citep{Elsayed2022SAViTE} improves SAVi by incorporating depth information, enabling the modeling of static scenes with camera motion.
OCLR~\cite{xie2022segmenting} adopts layered optical flow representations to identify multiple objects across the video.  However, OCLR faces challenges in maintaining long-term temporal consistency when applied to real-world scenes.

\myparagraph{Unsupervised 3D scene decomposition.}
Recent methods~\citep{kabra2021simone,chen2021roots,stelzner2021decomposing,yu2021unsupervised,sajjadi2022object} combine object-centric representations with view-dependent scene modeling techniques like neural radiance fields (NeRFs)~\citep{mildenhall2021nerf}.
ObSuRF~\citep{stelzner2021decomposing}
adopts the spatial broadcast decoder and takes depth information as training supervision.
uORF~\citep{yu2021unsupervised} extracts the background latent and foreground latents from an input static image to handle background and foreground objects separately.
%
%
For dynamic scenes, Guan \textit{et al.}~\citep{guan2022neurofluid} proposed to use a set of particle-based explicit representations in the NeRF-based inverse rendering framework, which is particularly designed for fluid physics modeling.
Driess \textit{et al.}~\citep{2022-driess-compNerfPreprint} explored the combination of an object-centric auto-encoder and volume rendering for dynamic scenes, which is relevant to our work. However, different from our unsupervised learning approach, it requires pre-prepared 2D object segments.
MovingParts~\cite{yang2024movingparts} considers motion as an important cue for identifying parts and parameterizes the scene motion by tracking the trajectory of particles on objects under a Lagrangian view, facilitating part-based representation learning.
%

\myparagraph{NeRF-based dynamic scene rendering.}
There is another line of work that models 3D dynamics using NeRF-based methods \citep{Pumarola2020DNeRFNR, li2021neural, Liu2022DeVRFFD, Wu2022D2NeRFSD, Guo_2022_NDVG_ACCV, li2023dynibar}.
D-NeRF~\citep{Pumarola2020DNeRFNR} uses a deformation network to map the coordinates of the dynamic fields to the canonical space. 
%
Li \textit{et al.}~\citep{li2021neural} extended the original MLP in NeRF to incorporate the dynamics information and determine the 3D correspondence of the sampling points at nearby time steps.
Li \textit{et al.}~\citep{li2023dynibar} achieved significant improvements on dynamic scene benchmarks by representing motion trajectories by finding 3D correspondences for sampling points in nearby views.
DeVRF~\citep{Liu2022DeVRFFD} models dynamic scenes with volume grid features~\citep{sun2022direct} and voxel deformation fields. 
$\text{D}^2$NeRF~\citep{Wu2022D2NeRFSD} presents a motion decoupling framework. 
Unlike our approach, it cannot segment multiple moving objects.
NeuralDiff~\citep{tschernezki2021neuraldiff} extends the standard NeRF reconstruction of static scenes with two dynamic components: one for transient objects and another for actors in egocentric videos. 
HyperNeRF~\citep{park2021hypernerf} extends the canonical space to a higher dimension to reconstruct topologically varying scenes with discontinuous deformations. 
TineuVox~\cite{fang2022fast} proposes to use explicit voxel grids to model temporal information, accelerating the learning time for dynamic scenes. However, these methods do not achieve an object-centric representation for dynamic scenes.

%% file: text/concl.tex
\section{Conclusion}

In this paper, we presented \newModel{}, an inverse graphics method designed to understand 3D dynamic scenes using object-centric volumetric representations. Our approach introduces two novel contributions:
First, it introduces a set of object-centric voxel grids to represent 3D dynamic scenes, ensuring a view-consistent understanding of object geometries and physical interactions. 
Second, it integrates 2D semantic priors learned from large-scale image datasets into 3D rendering through the volume slot attention module, significantly enhancing the model's ability to capture complex real-world scenes.
\newModel{} demonstrates remarkable performance compared to existing models in both unsupervised scene decomposition and novel view synthesis across synthetic and real-world scenarios. Additionally, our approach extends beyond existing neural rendering techniques for dynamic scenes by offering additional capabilities for dynamic scene editing.

%% file: dynavol_suppl.tex
\clearpage
\appendix
\IEEEpeerreviewmaketitle

\begin{figure*}[b]
\begin{center}
\includegraphics[width=0.99\linewidth]{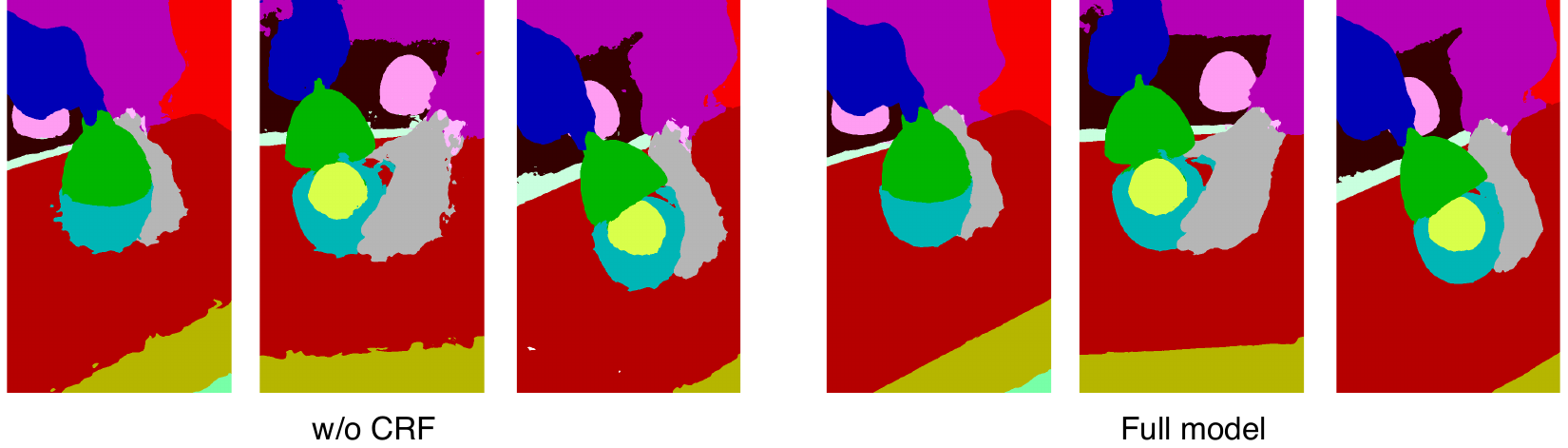}
\vspace{-5pt}
\caption{Comparisons of the scene decomposition results without and with the CRF post-processing method.
\label{fig:crf_compare}}
\end{center}
\end{figure*}

\section*{1 \quad Analysis of CRF Post-Processing} 
In Fig.~\ref{fig:crf_compare}, we compare the scene decomposition results without and with the proposed CRF post-processing. It can be found that CRF effectively mitigates the potential noise introduced by the initialization of the object-centric voxel grids. The final model provides more smooth 3D decomposition results, particularly for the object boundaries.

\section*{2 \quad \major{Analysis of handling complex shapes}}
\major{Achieving accurate 3D decomposition results for objects with complex shapes is especially challenging when the view scope is limited, as demonstrated in real-world scenes like the \textit{leaves} example in Fig.~\ref{fig:seg_realworld2} in the manuscript.}
\major{To address this issue, we originally use ``\textit{DINOv2-S with FeatUp}'' to provide semantic priors, which generates higher-resolution feature maps (\textit{i.e.}, $512 \times 512$) from $480 \times 270$ inputs. Although the high-resolution semantic maps enable fine-grained scene understanding, the FeatUp method can introduce blurring effects that may degrade the 3D scene decomposition results for objects with intricate shapes.}

\begin{figure*}[h]
\begin{center}
\includegraphics[width=0.99\linewidth]{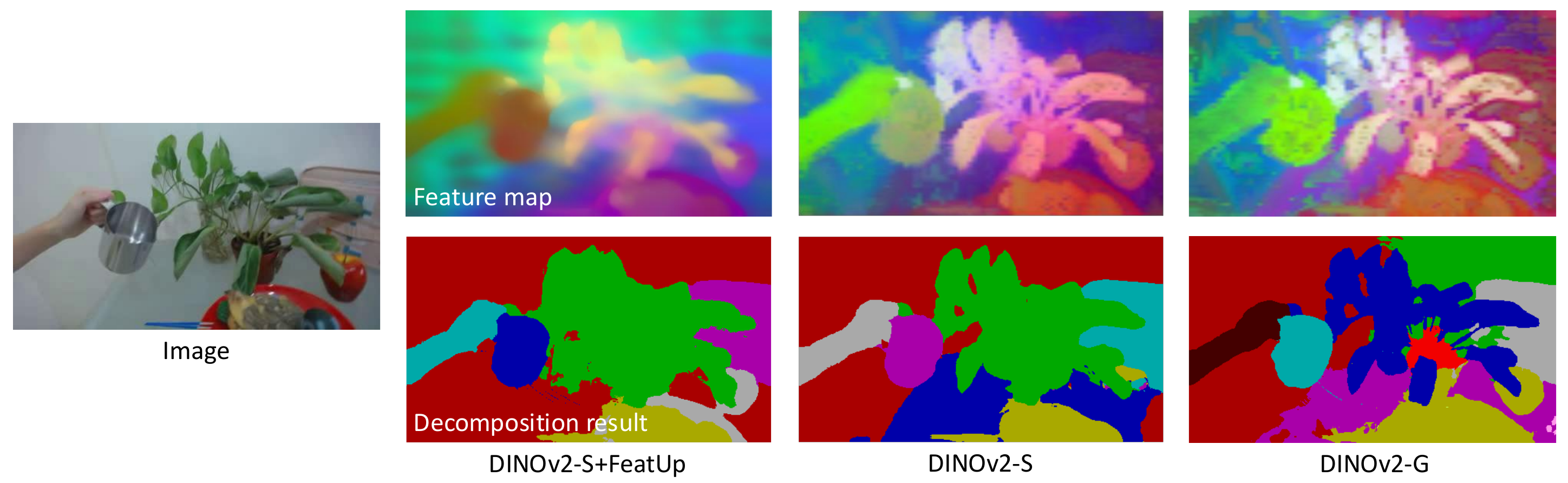}
\vspace{-5pt}
\caption{\major{Comparisons of the feature map and corresponding scene decomposition result with different semantic models.}
\label{fig:dino_giant}}
\end{center}
\end{figure*}

\major{To cope with complex objects such as the leaves, we remove the FeatUp method and experiment with two alternative semantic extractors, including ``\textit{DINOv2-S w/o FeatUp}'' and a more powerful model, ``\textit{DINOv2-G}''.}
\major{As shown in Fig.~\ref{fig:dino_giant}, both models produce clearer feature maps and improve scene decomposition results. }
\major{However, without FeatUp, we must resize the input images to larger dimensions to maintain the output semantic map at an acceptable resolution, \textit{i.e.}, $80 \times 80$. This resizing introduces noisy semantic maps, leading to over-segmentation of the 3D objects, especially with DINOv2-G, as these models were pre-trained on significantly smaller input sizes.}

\major{Consequently, an ideal solution would be to use ``\textit{DINOv2-G with FeatUp}'' to obtain both high-resolution and high-quality semantic maps. However, this model is not publicly available. We believe that this combination could significantly enhance the performance of our model in capturing fine-grained details, such as those seen in complex shapes like leaves, and represents a promising direction for future research.}


\section*{3 \quad More Real-World Examples}

\begin{figure*}[t]
\begin{center}
\includegraphics[width=\linewidth]{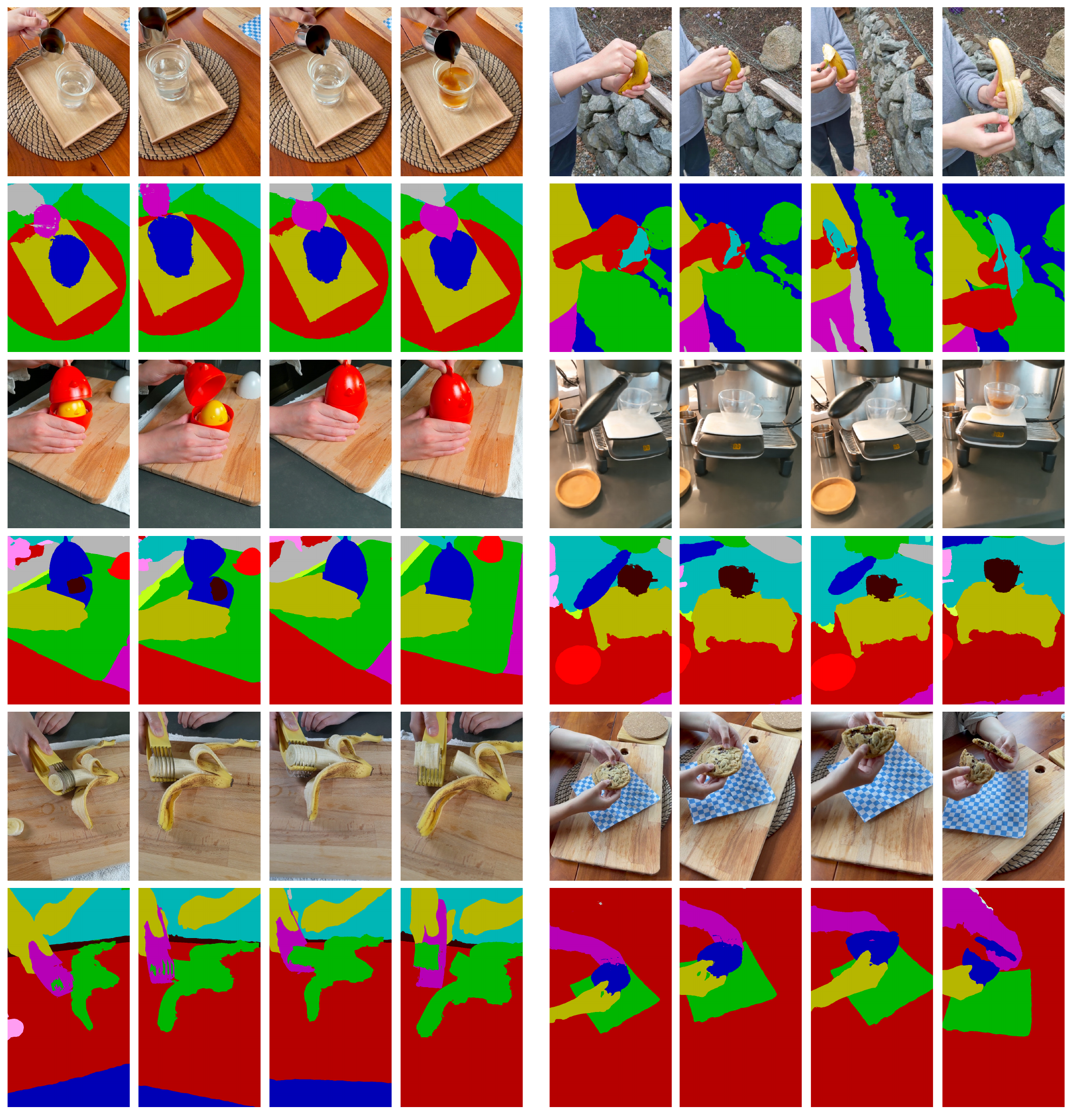}
\vspace{-5pt}
\caption{Additional examples of real-world decomposition results (Part 1).
\label{fig:realworld_supp_1}}
\end{center}
\end{figure*}

\begin{figure*}[t]
\begin{center}
\includegraphics[width=\linewidth]{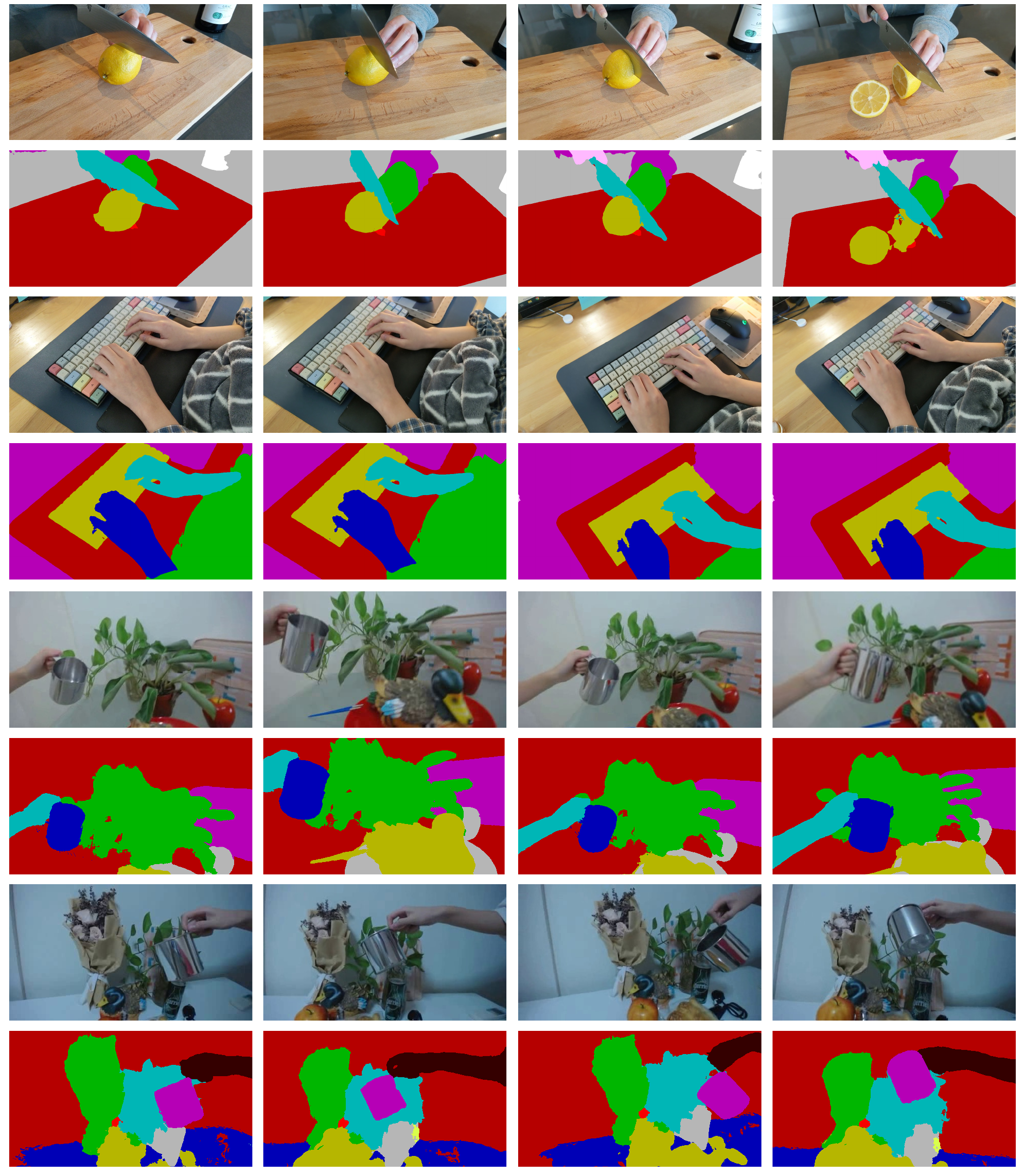}
\vspace{-5pt}
\caption{Additional examples of real-world decomposition results (Part 2).
\label{fig:realworld_supp_1_part2}}
\end{center}
\end{figure*}

\begin{figure*}[t]
\begin{center}
\includegraphics[width=\linewidth]{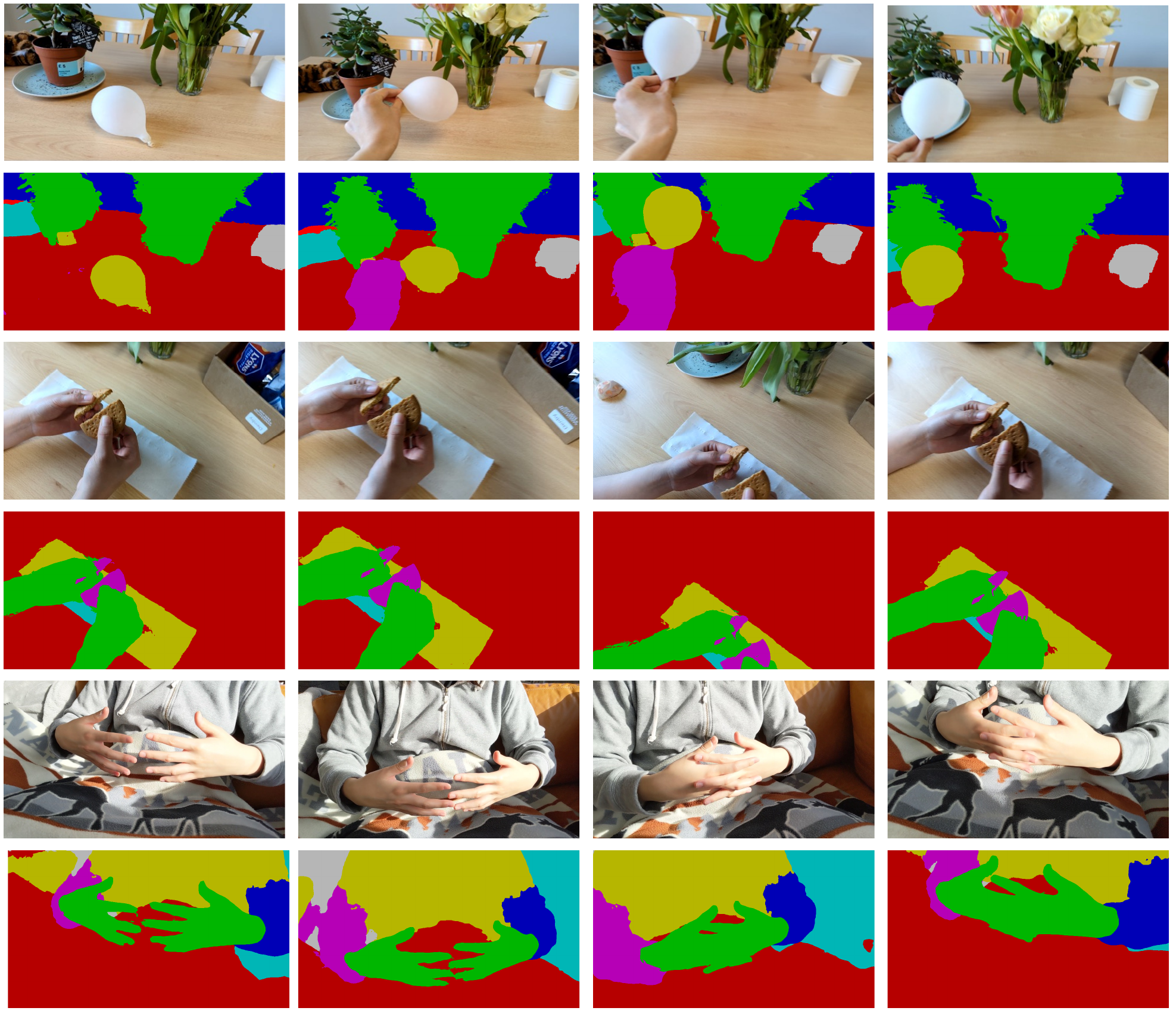}
\vspace{-5pt}
\caption{Additional examples of real-world decomposition results (Part 3).
\label{fig:realworld_supp_1_part3}}
\end{center}
\end{figure*}

In Fig. \ref{fig:realworld_supp_1}--\ref{fig:realworld_supp_1_part3}, we showcase additional real-world examples that are adopted from HyperNeRF~\cite{park2021hypernerf}, $\text{D}^2$NeRF~\citep{Wu2022D2NeRFSD}, and NeRF-DS~\citep{zhiwen2023nerfds}.
\newModel{} demonstrates impressive 3D decomposition results, even for scenes with complex object interactions and geometries.
For more examples of real-world scenes, please visit our project page: \url{https://zyp123494.github.io/DynaVol-S.github.io/}.

\section*{4 \quad Per-Object Rendering Results} 

\begin{figure*}[t]
\begin{center}
\includegraphics[width=\linewidth]{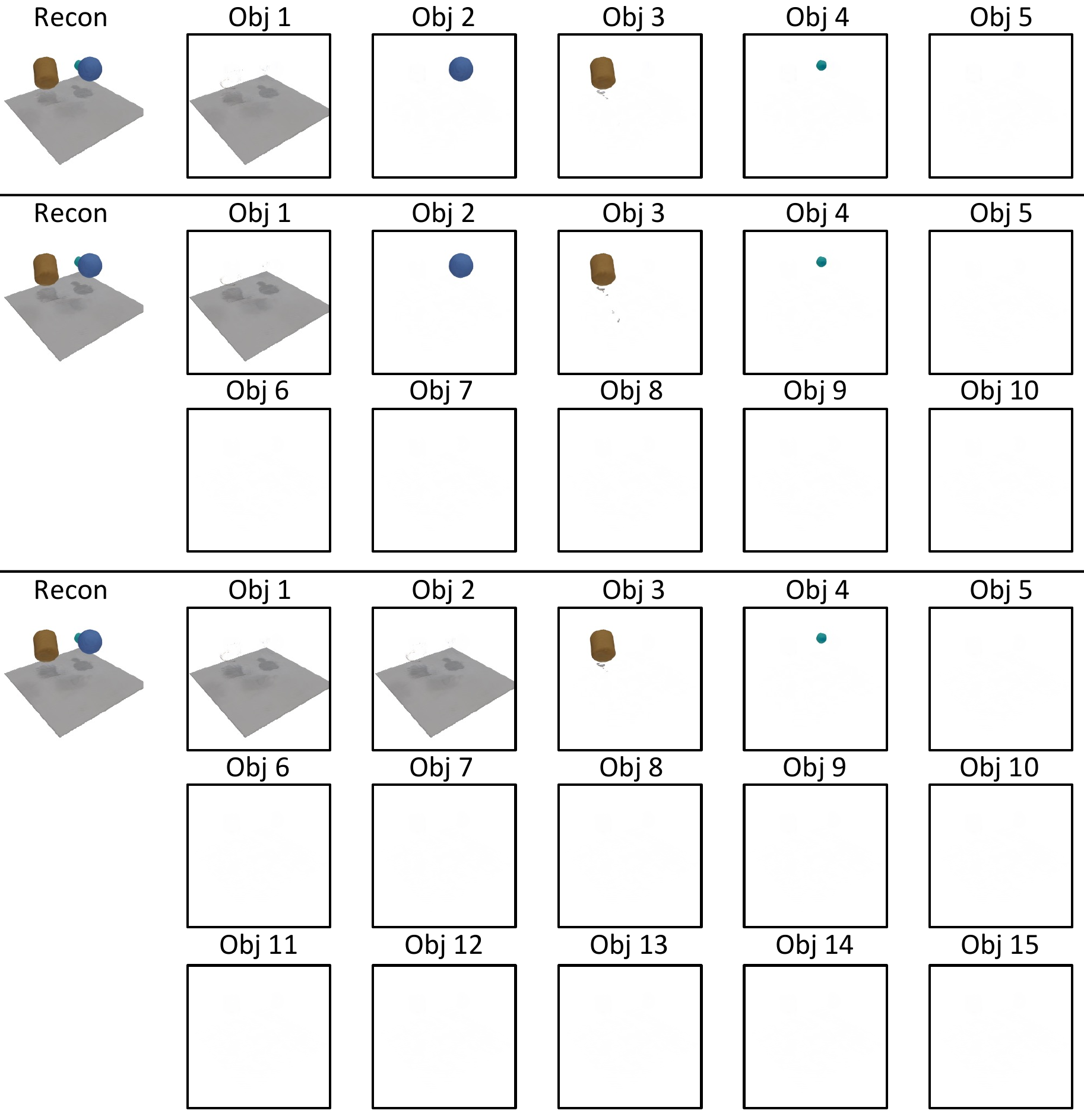}
\vspace{-5pt}
\caption{Per-object rendering results for scene \textit{3ObjRand} with $N=5$, $N=10$, and $N=15$, respectively.
\label{fig:per_slot}}
\end{center}
\end{figure*}

Fig.~\ref{fig:per_slot} showcases the per-object rendering results and demonstrates the robustness of our approach to the number of pre-defined latent codes or ``object slots''. 
It reveals that using a larger number of latent codes will not lead to the over-segmentation problem, as the redundant latent codes typically remain unassigned and do not contribute to the final rendering results.

\begin{figure*}[htbp]
\begin{center}
\centerline{
\includegraphics[width=\linewidth]{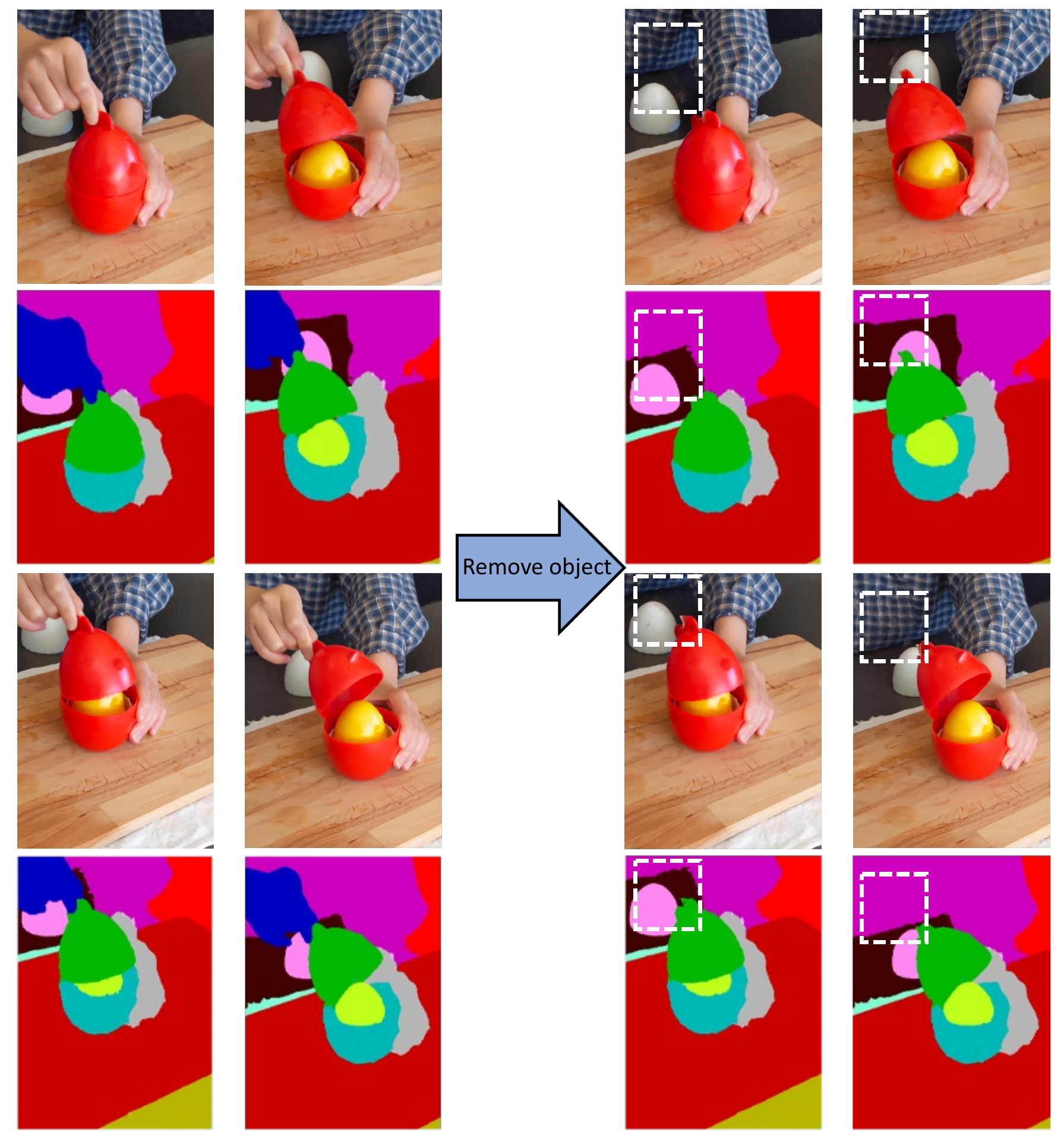}}
\caption{Real-world decomposition and scene editing results. The left images illustrate the original scene before editing, while the right images present the scene after object removal. It is an enlarged version of Fig. 7 in the main manuscript.
}
\label{fig:removal_chicken_supp}
\end{center}
\end{figure*}

\begin{figure*}[htbp]
\begin{center}
\centerline{
\includegraphics[width=0.9\linewidth]{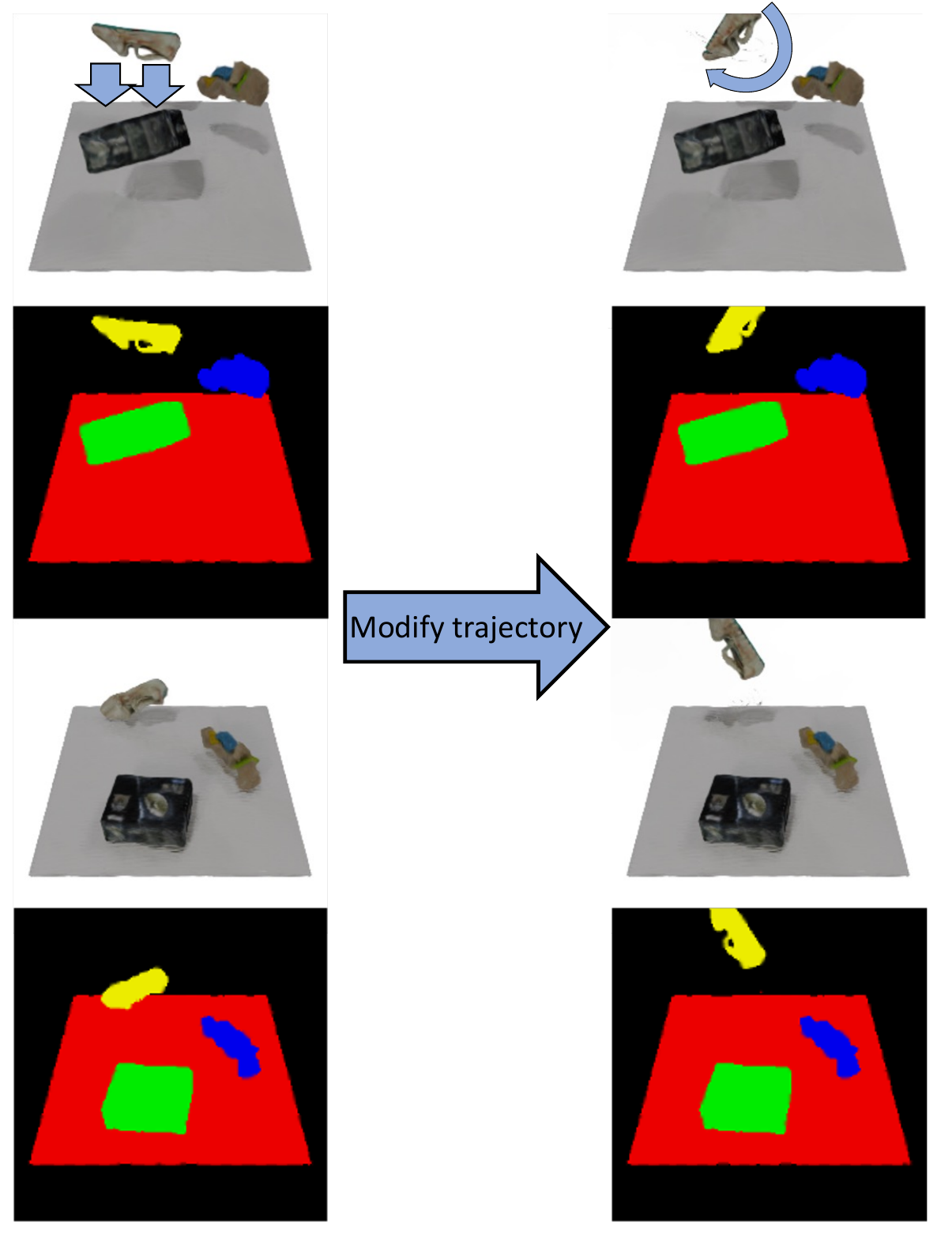}}
\caption{Trajectory modification based on \textit{3ObjRealCmpx} (Left: before editing; Right: after editing). It is an enlarged version of Fig. 7 in the main manuscript.
}
\label{fig:editing_cmp_supp}
\end{center}
\end{figure*}

\begin{figure*}[htbp]
\begin{center}
\centerline{
\includegraphics[width=0.9\linewidth]{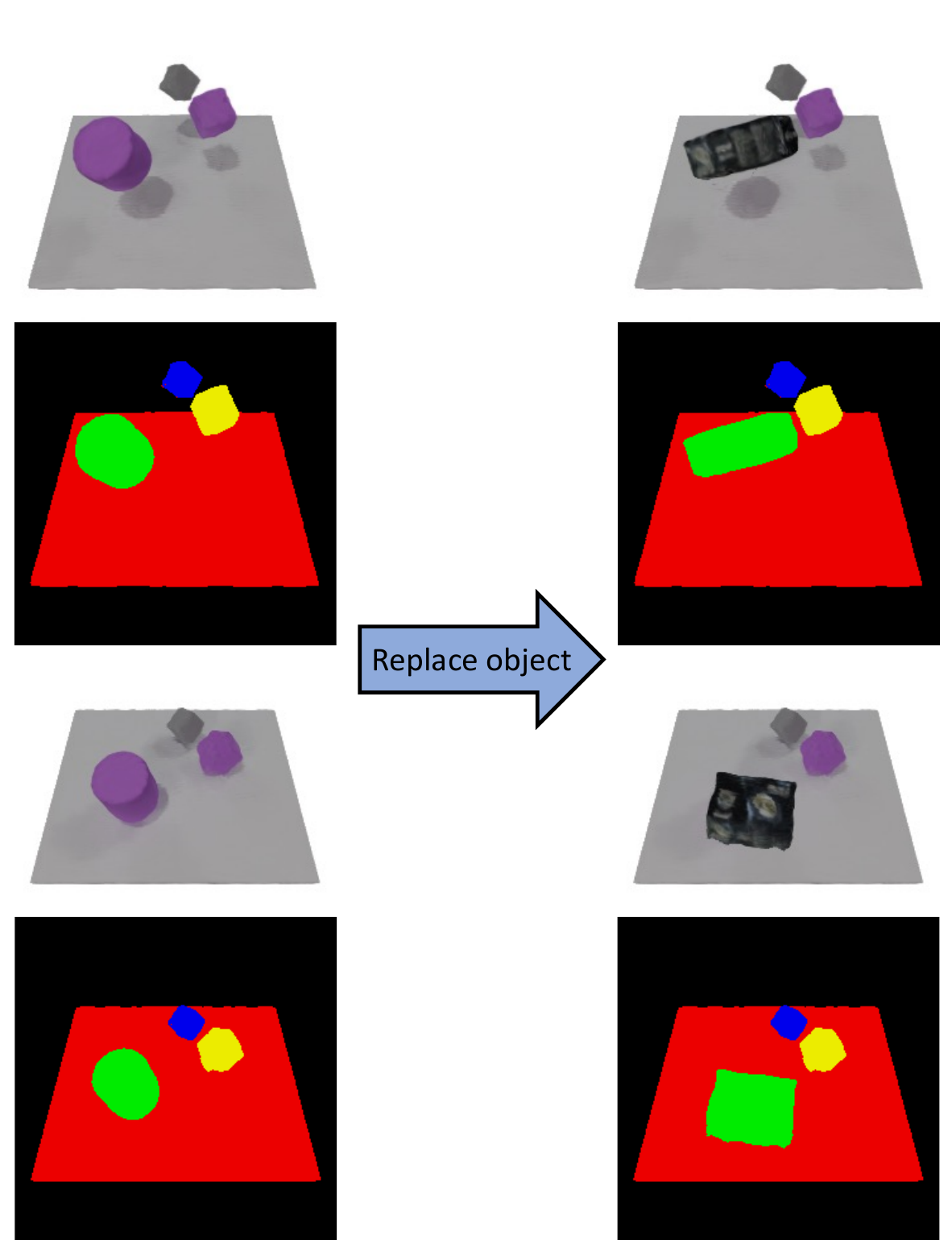}}
\caption{Object replacement based on \textit{3ObjFall} and \textit{3ObjRealCmpx} (Left: before editing; Right: after editing). 
}
\label{fig:replace_supp}
\end{center}
\end{figure*}

\begin{figure*}[htbp]
\begin{center}
\centerline{
\includegraphics[width=0.9\linewidth]{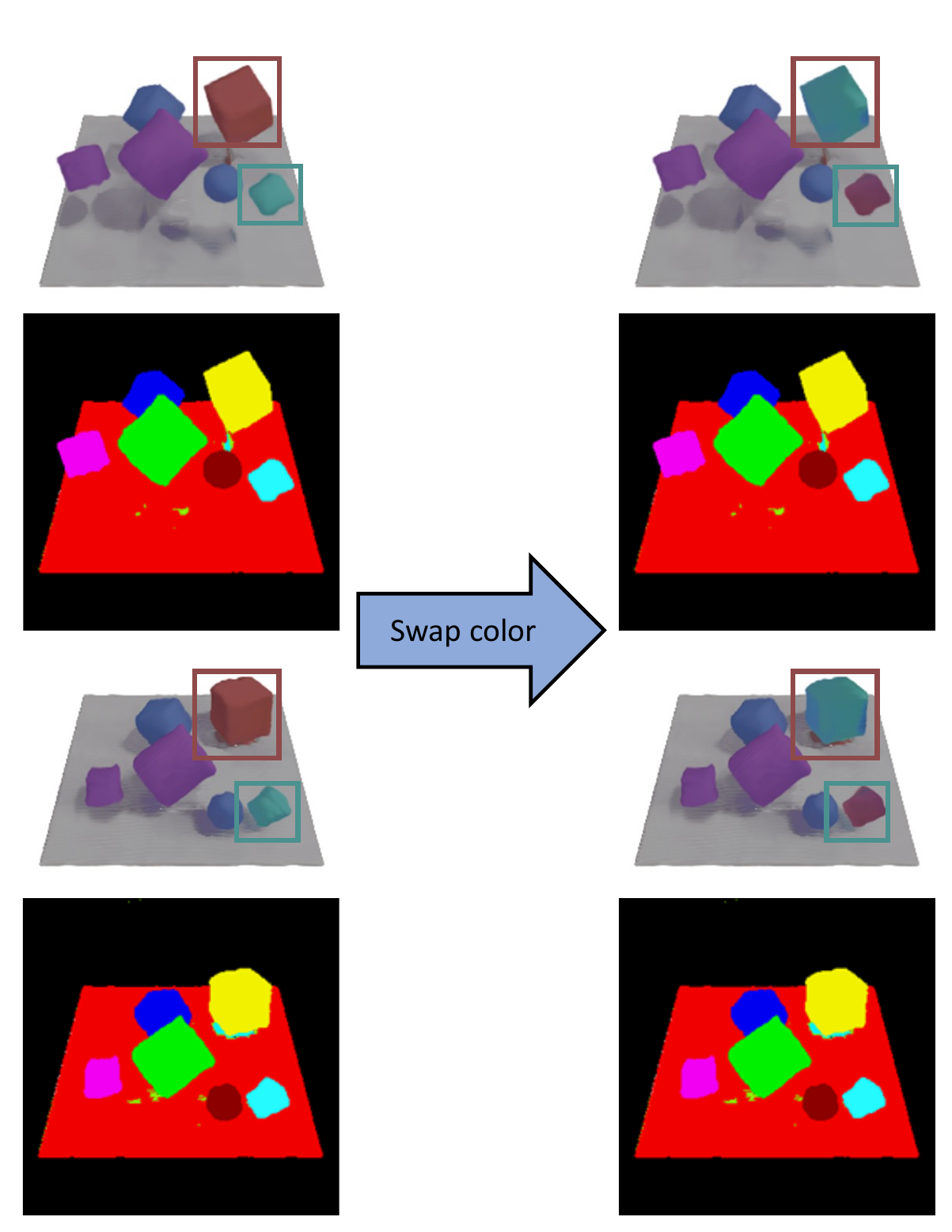}}
\caption{Color swapping based on \textit{6ObjFall} (Left: before editing; Right: after editing).
}
\label{fig:swap_supp}
\end{center}
\end{figure*}

\section*{5 \quad More Scene Editing Examples}
In Fig. \ref{fig:removal_chicken_supp}--\ref{fig:swap_supp}, we provide more showcases of dynamic scene editing obtained from \newModel{}. 
These figures provide a closer look at specific changes (\textit{e.g.}, object removal, motion modification, object replacement, and color swapping) made to the learned voxelized representations.
Please refer to the video demonstration on our project page for more examples.

\ifCLASSOPTIONcaptionsoff
  \newpage
\fi

